\documentclass[a4paper,fleqn]{cas-sc}

\renewcommand{\ttdefault}{lmtt}
\usepackage[numbers,sort&compress]{natbib}
\usepackage{etoolbox}
\AtBeginEnvironment{table}{\let\sffamily\rmfamily}
\AtBeginEnvironment{figure}{\let\sffamily\rmfamily}
\usepackage{amsmath,amssymb,amsthm}
\usepackage{algorithm,algorithmic}

\usepackage{float}
\usepackage{placeins}

\usepackage{stfloats}
\usepackage{colortbl}
\usepackage{multirow}
\usepackage{threeparttable}
\usepackage{tikz}
\usepackage{pgfplots}
\pgfplotsset{compat=1.18}
\usepackage{subcaption}
\usetikzlibrary{arrows.meta,positioning,fit,backgrounds,
                shapes.geometric,decorations.pathreplacing}

\theoremstyle{definition}
\newtheorem{definition}{Definition}
\newtheorem{example}{Example}
\theoremstyle{remark}
\newtheorem{remark}{Remark}
\theoremstyle{plain}
\newtheorem{theorem}{Theorem}

\newtheorem{proposition}{Proposition}

\definecolor{tabaggshade}{gray}{0.93}
\newcommand{\tabaggseparator}{\addlinespace[1pt]\midrule[\heavyrulewidth]\addlinespace[1pt]}
\newcommand{\tabaggrow}{\rowcolor{tabaggshade}\textbf{\textit{Mean over 11 datasets}}}
\newcommand{\fRIC}{Tol-RS$_\tau$}
\newcommand{\fRFS}{COFRS-min}
\newcommand{\Vobs}[1]{A_{#1}^{\mathrm{obs}}}
\newcommand{\VobsB}[1]{A_{#1,B}^{\mathrm{obs}}}
\newcommand{\RCO}{R_{\mathrm{CO}}}

\newcommand{\CDunit}{1.12}
\newcommand{\CDlabelL}{-2.55}
\newcommand{\cdCDscale}[2]{
  \pgfmathsetmacro{\cdlen}{#2*\CDunit}
  \draw[gray!70, line width=0.9pt, <->]
    (0,0.95) -- (\cdlen,0.95)
    node[midway, above=2pt, font=\footnotesize] {CD$=#1$};
}
\newcommand{\cdAxis}[2]{
  \pgfmathsetmacro{\xmin}{(#1-1)*\CDunit}
  \pgfmathsetmacro{\xmax}{(#2-1)*\CDunit}
  \pgfmathsetmacro{\CDaxisEnd}{\xmax}
  \pgfmathsetmacro{\CDaxisMid}{0.5*(\xmin+\xmax)}
  \draw[thick] (\xmin,0) -- (\xmax,0);
  \foreach \r in {1,2,3,4,5,6} {
    \pgfmathtruncatemacro{\ri}{\r}
    \ifnum\ri<#1\relax\else
      \ifnum\ri>#2\relax\else
        \pgfmathsetmacro{\x}{(\r-1)*\CDunit}
        \draw (\x,0.07) -- (\x,-0.07);
        \node[above=3pt, font=\footnotesize] at (\x,0) {\r};
      \fi
    \fi
  }
}
\newcommand{\cdLeft}[3]{
  \pgfmathsetmacro{\xr}{(#1-1)*\CDunit}
  \pgfmathsetmacro{\y}{-0.40*#3}
  \draw (\xr,0) -- (\xr,\y) -- (\CDlabelL,\y);
  \node[anchor=east, font=\normalsize] at (\CDlabelL,\y) {#2};
}
\newcommand{\cdRight}[4]{
  \pgfmathsetmacro{\xr}{(#1-1)*\CDunit + #4}
  \pgfmathsetmacro{\y}{-0.40*#3}
  \pgfmathsetmacro{\xR}{\CDaxisEnd+2.05}
  \draw (\xr,0) -- (\xr,\y) -- (\xR,\y);
  \node[anchor=west, font=\normalsize] at (\xR,\y) {#2};
}
\newcommand{\cdClique}[4]{
  \pgfmathsetmacro{\xa}{(#1-1)*\CDunit}
  \pgfmathsetmacro{\xb}{(#2-1)*\CDunit + #4}
  \draw[line width=2.2pt] (\xa,#3) -- (\xb,#3);
}
\newcommand{\cdTitle}[1]{
  \node[font=\normalsize\bfseries] at (\CDaxisMid,1.35) {#1};
}

\hypersetup{colorlinks=true,allcolors=blue,hypertexnames=false,
  pdftitle={Conservative Overlap Fuzzy Rough Sets for Incomplete Data},
  pdfkeywords={Rough sets; Fuzzy rough sets; Attribute reduction; Incomplete data; Overlap similarity; Mixed attributes}}

\usepackage{pdfpages}
\ExplSyntaxOn
\cs_set:Npn \__first_footerline: { \group_begin: \small \sffamily \__short_authors: :~ { \rmfamily \itshape Preprint } \group_end: }
\ExplSyntaxOff
\begin{document}
\let\WriteBookmarks\relax
\raggedbottom
\shorttitle{Conservative overlap fuzzy rough sets}
\shortauthors{T.H. Vuong et al.}

\title[mode = title]{Conservative Overlap Fuzzy Rough Sets for Incomplete Data}

\author[1,3]{Trung Hieu Vuong}
\author[3]{Van Thien Nguyen}
\author[2]{Nhu Son Nguyen}
\author[4]{Thanh Dai Tran}
\cormark[1]
\ead{ttdaiuneti@gmail.com}

\affiliation[1]{organization={Graduate University of Science and Technology, Vietnam Academy of Science and Technology},
  addressline={18 Hoang Quoc Viet Road}, city={Hanoi}, country={Vietnam}}
\affiliation[2]{organization={Institute of Information Technology, Vietnam Academy of Science and Technology},
  addressline={18 Hoang Quoc Viet Road}, city={Hanoi}, country={Vietnam}}
\affiliation[3]{organization={Hanoi University of Industry},
  addressline={298 Cau Dien Street}, city={Hanoi}, country={Vietnam}}
\affiliation[4]{organization={University of Economics - Technology for Industries},
  addressline={454--456 Minh Khai Street}, city={Hanoi}, country={Vietnam}}
\cortext[cor1]{Corresponding author.}

\begin{abstract}
Missing values break the equivalence classes on which rough set attribute reduction relies.
The usual tolerance relation lets a missing value match any value, so objects with many missing
values look alike.
We propose the Conservative Overlap Fuzzy Rough Set (COFRS) model. Its similarity is the weighted
overlap of the attributes present in both objects, divided by the fixed number of attributes $|A|$,
so every missing value lowers the similarity.
From this similarity we define a soft dependency degree on $\delta$-granules and search for reducts
by greedy backward elimination in $O(|U|^2\cdot|A|^2)$ time.
We show that the relation is symmetric and not reflexive and that the dependency degree is not
monotone, and we give a sufficient condition for monotonicity.
On 11 incomplete UCI benchmarks the reduct size is set largely by the granule threshold.
At equal size, the attributes chosen by COFRS give a higher $k$NN G-mean than random subsets and
than the exact-match tolerance ranking, but not than the other rankings.
COFRS returns smaller reducts than two of the three tolerance-based methods, with no significant
difference in $k$NN accuracy, and it has the weakest results with an RBF-SVM.
The gain is compactness, not predictive accuracy.
\end{abstract}


\maketitle
\thispagestyle{empty}

\section{Introduction}
\label{sec:intro}

In real-world information systems, missing values are common because of sensor
failures, privacy constraints, or incomplete data collection~\cite{Garcia2010,Sterne2009}.
Attribute reduction identifies a small subset of attributes that keeps the discriminative
power of the original data, and it is a basic step in machine learning and data
mining~\cite{Chandrashekar2014,Li2017}.
Rough set theory (RST)~\cite{Pawlak1982,Pawlak1991} offers a rigorous framework for it that
uses only the data and needs no prior domain knowledge.

Missing values break the basic construction of RST. A granule is an equivalence class, and
equivalence requires every compared value to be present.
The usual remedy is a tolerance relation in which a missing value matches any value~\cite{Kryszkiewicz1998}.
This optimistic rule makes objects look alike exactly when little is known about them.
An object with many missing values is then tolerant to many others, its class mixes decisions,
and the dependency degree $\gamma$ of small subsets drops. More attributes are needed to separate
the classes, so the reducts grow with the missing rate.

Existing work on this problem falls into three groups.
Tolerance-based methods are the classical line for incomplete systems.
Kryszkiewicz~\cite{Kryszkiewicz1998} introduced tolerance relations and discernibility-based
reduction, and Stefanowski and Tsoukias~\cite{Stefanowski1999} extended rough sets under
incomplete information.
Later work speeds up the computation~\cite{Meng2009,Qian2011} or makes it
incremental~\cite{Cheng2023}.
The relation of Kryszkiewicz and the methods built on it keep the binary wildcard semantics,
and therefore its inflation of similarity under high missing rates.

Fuzzy rough sets~\cite{DuboisPrade1990,Radzikowska2002} replace the crisp relation by a
graded one, and Yuan et al.~\cite{Yuan2021} review and compare the reduction variants.
Jensen and Shen~\cite{Jensen2004} select attributes with the fuzzy positive region, and later
work~\cite{Jensen2009} adds selectors based on the fuzzy boundary region and a fuzzy discernibility matrix.
Tran et al.~\cite{Tran2024IFT} define the reduct of a decision table through an intuitionistic fuzzy topology.
Kernelised and fitting-model extensions~\cite{Hu2011,Wang2017} also target feature
selection, but they typically assume complete or preprocessed data.
He et al.~\cite{He2022} proposed a fuzzy rough reduction for incomplete \emph{categorical}
decision systems; it does not handle continuous or mixed attributes without discretisation.
Fuzzy rough sets have also been used to impute missing values~\cite{Amiri2016Imputation}.
Lenz et al.~\cite{Lenz2021} model a missing value with an optimistic and a pessimistic
per-attribute relation (the value counts as $1$ or as $0$) and average over the attributes.
With $\mu_a$ as the per-attribute relation, their pessimistic relation coincides with our $\RCO$ on the full attribute set.
They use it for nearest-neighbour classification and do not study reduct search.
These methods need complete data, are limited to categorical attributes, or do not search for reducts.

A third group targets mixed incomplete data with finer similarity models.
Zhao and Qin~\cite{Zhao2014} define a neighbourhood-tolerance relation for mixed categorical and
numerical attributes and select attributes by a conditional entropy; the relation is binary and a
missing value matches any value.
Zhou and Bao~\cite{Zhou2023} use a binary similarity matrix.
Wu et al.~\cite{Wu2024} relate two objects only when their neighbourhood-tolerance classes
coincide. This avoids the extra threshold of earlier tolerance refinements, but the relation keeps
a neighbourhood radius and stays binary.
Li et al.~\cite{Li2023} cover incomplete mixed data with neighbourhoods and select attributes by a
max-relevance, min-redundancy uncertainty function, and Li et al.~\cite{Li2021Hetero} select
attributes of mixed data with a fuzzy information entropy.
Nguyen et al.~\cite{Nguyen2026MHMG} remove the radius parameter by setting the granule radius of each object to its distance to the nearest object of another class. The method works on numeric data and does not treat missing values.
Tang and Qiao~\cite{Tang2024} combine three-way decisions with triangular norms, and
Dai~\cite{Dai2013,Dai2024} and Wang et al.~\cite{Wang2019} study numerical and
distance-based fuzzy rough reductions.
As far as we can tell, none of these works models the overlap of present-attribute sets under missingness.

Two recent reductions also target incomplete data.
Chen et al.~\cite{Chen2025} fuse an incomplete Euclidean distance with a global similarity to relate
objects with missing values. They add an incomplete cosine similarity between conditional
attributes and select attributes by the resulting bidirectional fuzzy similarity discriminability.
Zhang et al.~\cite{Zhang2025} define a fuzzy granularity entropy for incomplete real-valued decision
systems, show that its conditional form is monotone, and update the reduct incrementally when objects
change.
Other lines include large-scale local models~\cite{Li2024LFRS}, stability-driven reduct
optimisation~\cite{Gao2024RDROAR} and weighted-neighbourhood constrained models~\cite{He2024GWNCFRS}.
Xie et al.~\cite{Xie2025NOD} define an overlap degree between attributes and use it to pre-sort
attributes for fuzzy rough reduction of real-valued data; their overlap relates attributes, whereas
ours relates the present-attribute sets of two objects.

Within the literature reviewed here, we are not aware of an attribute reduction method for
incomplete mixed data that combines an $|A|$-normalised overlap of present attributes with a
soft, granule-based dependency degree.
The nearest relation is the pessimistic one of Lenz et al.
They define it on the full attribute set only, so its direct restriction to a subset $B\subsetneq A$
averages over $|B|$, whereas $\RCO^B$ keeps the denominator $|A|$.

Our approach starts from a simple observation: a missing value is no evidence that two
objects are alike.
We therefore compare two objects only on the attributes present in both, and we divide the
matched similarity by the fixed number of attributes $|A|$, not by the number of compared
attributes.
Every missing value then lowers the similarity, whatever the other object holds.
This is a deliberate bias and not a model of the missingness mechanism.
It suits settings in which an absent value is itself informative, the
missing-not-at-random case in Rubin's taxonomy~\cite{Rubin1976}.
We develop the idea into a fuzzy rough set model, COFRS, and make three
contributions.

1)~We define the conservative overlap similarity $\RCO$, a weighted overlap of present
attributes with the fixed denominator $|A|$, and a soft dependency degree $\gamma_{\mathrm{soft}}$
built from weighted $\delta$-granule consistency.
We show that $\RCO$ is symmetric and non-reflexive and that $\gamma_{\mathrm{soft}}$ is not
monotone in the attribute set. Granule stability and purity alignment together are
sufficient for monotonicity.

2)~We give a greedy backward elimination algorithm that updates $\RCO$ incrementally and runs
in $O(|U|^2\cdot|A|^2)$ time.

3)~We compare COFRS on 11 incomplete UCI datasets~\cite{UCI2023} with three tolerance-based
methods, an ablation that aggregates the attribute similarities by $\min$, and two further
fuzzy-rough and neighbourhood selectors.
When the reduct size is held fixed, the attributes of COFRS give a better $k$NN G-mean than random
subsets of that size and than the top attributes ranked by Tol-RS. Its reducts are smaller than those of two of the three tolerance-based
methods, and its $k$NN accuracy does not differ significantly from theirs.
A sensitivity study of $\delta$ and an injected-missingness experiment show that $\delta$ sets the
reduct size and that the penalty limits the reduct growth on numeric but not on nominal data.

The gain of COFRS is compactness, not accuracy. Against Tolerance-RS the size difference is not
significant, COFRS has the weakest SVM results, and the conservative bias does not suit every
missingness mechanism.

The remaining sections build the three results in the order in which they depend on each other.
Section~\ref{sec:prelim} first recalls the rough set notions and the tolerance relation that
COFRS replaces, so that the change is visible.
The overlap similarity, the soft dependency and the properties on which the reduct search
relies come next, in Section~\ref{sec:model}, and Section~\ref{sec:algorithm} turns them into the
backward search and states its cost.
The experiments of Section~\ref{sec:experiments} then test the reducts that this search returns,
and Section~\ref{sec:discussion} uses the theory to explain where they help and where they do not.
Section~\ref{sec:conclusion} concludes.

\section{Preliminaries}
\label{sec:prelim}

This section recalls the notions on which COFRS is built: information systems and reducts, the
tolerance relation that classical rough sets use for missing values, and fuzzy rough sets.
Example~\ref{ex:prelim} shows how the tolerance relation inflates similarity. Section~\ref{sec:model}
replaces it by a graded overlap similarity, and Section~\ref{sec:algorithm} uses the resulting
dependency degree to search for reducts.

\subsection{Information Systems and Attribute Reduction}
\label{subsec:is}

\begin{definition}[Information system~\cite{Pawlak1991}]
\label{def:is}
An information system (IS) is a tuple $IS=(U,A,V,f)$, where
$U=\{x_1,\ldots,x_n\}$ is a finite set of objects, $A=\{a_1,\ldots,a_m\}$ is a finite set of
attributes, $V=\bigcup_{a\in A} V_a$ is the union of the attribute domains $V_a$, and
$f:U\times A \rightarrow V$ is an information function with $f(x,a)\in V_a$.
\end{definition}

\begin{definition}[Decision table~\cite{Pawlak1991}]
\label{def:ds}
A decision table is an IS together with a decision attribute $D\notin A$,
$D:U\rightarrow V_D$. The decision classes are $D_d=\{x\in U: D(x)=d\}$ for $d\in V_D$.
The decision attribute is assumed to be complete.
\end{definition}

\begin{definition}[Incomplete information system~\cite{Kryszkiewicz1998}]
\label{def:iis}
An IS is incomplete if some entries are missing.
We denote a missing value by $*$ and write
$IS=(U,A,V\cup\{*\},f)$, where $f(x,a)=*$ when the value of $a$ for $x$ is unknown.
\end{definition}

\begin{definition}[Present attribute set]
\label{def:vx}
For each $x\in U$, the present attribute set is
\begin{equation}
\Vobs{x} = \{a\in A : f(x,a)\neq *\}.
\label{eq:vx}
\end{equation}
\end{definition}

\subsection{Classical Rough Set and Tolerance Relation}
\label{subsec:tolerance}

\begin{definition}[Indiscernibility and granule~\cite{Pawlak1982}]
\label{def:indisc}
Let the IS be complete and $B\subseteq A$. The indiscernibility relation is
$R_B(x,y)=1$ iff $f(x,a)=f(y,a)$ for all $a\in B$. It is an equivalence relation, and the
granule of $x$ is $[x]_B=\{y\in U : R_B(x,y)=1\}$.
\end{definition}

\begin{definition}[Positive region and dependency degree~\cite{Pawlak1982}]
\label{def:gamma}
For a decision table and $B\subseteq A$,
\begin{equation}
\mathrm{POS}_B(D)=\{x\in U : [x]_B\subseteq D_{D(x)}\},\qquad
\gamma_B(D)=\frac{|\mathrm{POS}_B(D)|}{|U|}.
\label{eq:dependency}
\end{equation}
\end{definition}

\begin{definition}[Classical reduct~\cite{Pawlak1991}]
\label{def:reduct}
A subset $B\subseteq A$ is a reduct if $\gamma_B(D)=\gamma_A(D)$ and
no proper subset $B'\subset B$ satisfies $\gamma_{B'}(D)=\gamma_A(D)$.
\end{definition}

\begin{definition}[Tolerance relation~\cite{Kryszkiewicz1998}]
\label{def:tolerance}
For an incomplete IS and $a\in A$, $T_a(x,y)=1$ if $f(x,a)=f(y,a)$, or $f(x,a)=*$, or
$f(y,a)=*$, and $T_a(x,y)=0$ otherwise:
\begin{equation}
T_a(x,y)=\mathbf{1}\bigl[\,f(x,a){=}f(y,a)\ \vee\ f(x,a){=}*\ \vee\ f(y,a){=}*\,\bigr].
\label{eq:tolerance}
\end{equation}
For $B\subseteq A$, $T_B(x,y)=\min_{a\in B} T_a(x,y)$, and the tolerance class of $x$ is
$T_B(x)=\{y\in U : T_B(x,y)=1\}$.
\end{definition}

For incomplete decision tables, $\mathrm{POS}_B(D)$ and $\gamma_B(D)$ are defined by
\eqref{eq:dependency} with $T_B(x)$ in place of $[x]_B$, and the reduct is defined as in
Definition~\ref{def:reduct}.
The tolerance relation is reflexive and symmetric but not transitive~\cite{Kryszkiewicz1998}.
It adopts an optimistic assumption: a missing value may match any observed value.
Under high missing rates this inflates similarity. Tolerance classes grow, $\gamma_B(D)$ drops,
and larger subsets $B$ are needed to separate the classes (Example~\ref{ex:prelim}).

\begin{example}
\label{ex:prelim}
Table~\ref{tab:toy_is} shows a decision table with four objects and three attributes.
For $B=A$ the tolerance classes are $T_A(x_1)=T_A(x_3)=\{x_1,x_3\}$, $T_A(x_2)=\{x_2\}$ and
$T_A(x_4)=\{x_4\}$. Each class lies inside one decision class, so
$\mathrm{POS}_A(D)=U$ and $\gamma_A(D)=1$.
The subset $B=\{a_1,a_2\}$ gives the same classes and $\gamma_B(D)=1$.
Its proper subsets do not: $T_{a_1}(x_1)=T_{a_1}(x_2)=T_{a_1}(x_3)=\{x_1,x_2,x_3\}$ contains
both labels, so $\gamma_{\{a_1\}}(D)=\tfrac14$, and likewise $\gamma_{\{a_2\}}(D)=\tfrac14$.
Hence $\{a_1,a_2\}$ is a reduct.
The wildcard effect is visible for $x_4$: its value of $a_2$ is missing, so
$T_{a_2}(x_4)=U$, although $x_4$ has no observed value of $a_2$ to match.
Example~\ref{ex:rco_gamma} reuses this table.
\end{example}

\begin{table}[pos=htbp]
\centering
\small
\caption{Toy incomplete decision table for Examples~\ref{ex:prelim} and~\ref{ex:rco_gamma}.}
\label{tab:toy_is}
\begin{tabular}{lcccc}
\toprule
Object & $a_1$ & $a_2$ & $a_3$ & $D$ \\
\midrule
$x_1$ & 1 & 1 & 1 & $+$ \\
$x_2$ & 1 & 0 & 1 & $-$ \\
$x_3$ & 1 & 1 & -- & $+$ \\
$x_4$ & 0 & -- & 1 & $-$ \\
\bottomrule
\end{tabular}
\end{table}

\subsection{Fuzzy Rough Sets}
\label{subsec:frs}

Fuzzy rough sets~\cite{DuboisPrade1990,Radzikowska2002} replace the crisp relation by a graded one.

\begin{definition}[Fuzzy relation~\cite{DuboisPrade1990}]
\label{def:fuzzy_relation}
A fuzzy relation on $U$ is a mapping $R:U\times U\rightarrow [0,1]$.
\end{definition}

\begin{definition}[Lower and upper approximation~\cite{DuboisPrade1990}]
\label{def:approx}
Given a fuzzy relation $R$ and $X\subseteq U$ with indicator $\chi_X$,
\begin{align}
\underline{R}(X)(x) &= \inf_{y\in U}\max\!\bigl(1-R(x,y),\,\chi_X(y)\bigr),
  \label{eq:lower_approx} \\
\overline{R}(X)(x) &= \sup_{y\in U}\min\!\bigl(R(x,y),\,\chi_X(y)\bigr).
  \label{eq:upper_approx}
\end{align}
\end{definition}

\begin{definition}[Fuzzy dependency degree~\cite{Jensen2004,Jensen2009}]
\label{def:fuzzy_gamma}
For a fuzzy relation $R_B$ induced by $B\subseteq A$, the fuzzy positive region and the
fuzzy dependency degree are
\begin{align}
\mathrm{POS}_B(D)(x) &= \max_{d\in V_D}\underline{R_B}(D_d)(x), \notag \\
\gamma_B(D) &= \frac{1}{|U|}\sum_{x\in U} \mathrm{POS}_B(D)(x).
\label{eq:fuzzy_dependency}
\end{align}
When $R_B$ is crisp, \eqref{eq:fuzzy_dependency} reduces to \eqref{eq:dependency}.
\end{definition}

\section{COFRS Model}
\label{sec:model}

This section defines the COFRS model and proves its properties.
We first state what the overlap similarity changes with respect to the tolerance relation and the
Jaccard index. We then define the similarity and prove its basic properties, and build the soft
dependency degree on it. The properties of the similarity lead to the non-monotonicity of the
dependency degree, and this is why the reduct search of Section~\ref{sec:algorithm} has to be tolerant.

\subsection{Motivation}
\label{subsec:motivation}

Tolerance relations let a missing value match any value (Section~\ref{sec:intro}), and
min-based fuzzy rough similarity~\cite{DuboisPrade1990} is defined for complete data.
Neither measures how much two objects overlap on the attributes present in both.
The Jaccard index~\cite{Jaccard1901} $|P\cap Q|/|P\cup Q|$ measures relative overlap, but its
union denominator shrinks when values are missing and so softens the penalty.
The proposed $\RCO$ fixes the denominator at $|A|$, so every missing attribute lowers the similarity.

\subsection{COFRS Framework}

\begin{figure}[pos=htbp]
\centering
\includegraphics[width=0.92\linewidth]{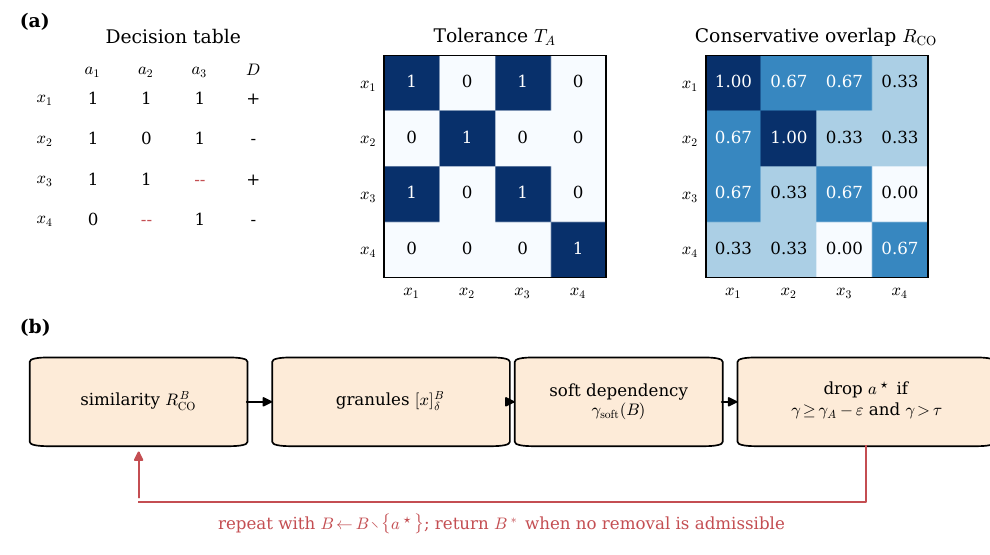}
\caption{COFRS on the toy table of Examples~\ref{ex:prelim} and~\ref{ex:rco_gamma}.
\textbf{(a)}~Tolerance relation and conservative overlap similarity.
\textbf{(b)}~Reduction loop: granules and $\gamma_{\mathrm{soft}}$ are recomputed for each candidate removal.}
\label{fig:pipeline}
\end{figure}

COFRS is built in three steps. We first define a per-attribute similarity $\mu_a$ for mixed data.
We then combine the attributes into the conservative overlap similarity $\RCO$, which is
$|A|$-normalised and penalises missing values. Finally, the $\delta$-granules of $\RCO$ give the
soft dependency degree $\gamma_{\mathrm{soft}}$. The reduction objective follows in
Section~\ref{subsec:finding_reduct}, next to the search algorithm.
Figure~\ref{fig:pipeline} illustrates the mechanism on a small table and the reduction loop.

\begin{definition}[Attribute Similarity]
\label{def:mu_a}
For attribute $a\in A$, the per-attribute similarity $\mu_a(x,y)\in[0,1]$ is defined as
\begin{equation}
\mu_a(x,y)=
\begin{cases}
\exp\!\left(-\tfrac{\bigl(f(x,a)-f(y,a)\bigr)^2}{\sigma_a^2}\right),
  & \text{$a$ continuous,}\\[4pt]
\mathbf{1}\bigl[f(x,a){=}f(y,a)\bigr],
  & \text{$a$ categorical,}
\end{cases}
\label{eq:mu_a}
\end{equation}
where $\sigma_a$ is the standard deviation of attribute $a$ on observed values
(missing entries excluded); for $\sigma_a=0$ use $\mathbf{1}[f(x,a){=}f(y,a)]$.
\end{definition}

\subsection{Conservative Overlap Similarity}
\label{subsec:rco}

\begin{definition}[Conservative overlap similarity]
\label{def:rco}
For $x,y\in U$, the conservative overlap (CO) similarity is the
$|A|$-normalised weighted overlap
\begin{equation}
\RCO(x,y)=\frac{\sum_{a\in \Vobs{x}\cap \Vobs{y}}\mu_a(x,y)}{|A|}.
\label{eq:rco}
\end{equation}
For $B\subseteq A$,
\begin{equation}
\RCO^B(x,y)=\frac{\sum_{a\in (\Vobs{x}\cap \Vobs{y})\cap B}\mu_a(x,y)}{|A|}.
\label{eq:rco_b}
\end{equation}
When $B=A$, $\RCO^A(x,y)=\RCO(x,y)$.
Equation~\eqref{eq:rco} is \textbf{not} the Jaccard index: the denominator is the fixed
schema size $|A|$, not the union $|\Vobs{x}\cup \Vobs{y}|$.
\end{definition}

The fixed denominator is the central design choice. A union denominator
$|\Vobs{x}\cup\Vobs{y}|$ gives a Jaccard-style overlap whose penalty shrinks when values are
missing, and $|B|$ in \eqref{eq:rco_b} would break the subset monotonicity of
Theorem~\ref{thm:monotonicity}. The two measures agree only when every attribute is present in at
least one of the two objects. For example, with $|A|=4$, $\Vobs{x}=\{a_1,a_2\}$,
$\Vobs{y}=\{a_1,a_3\}$ and $\mu_{a_1}(x,y)=1$, $\RCO(x,y)=1/4$ while the Jaccard index is $1/3$
(Section~S1 of the Supplementary material).

With $\mu_a$ as the per-attribute score, the sum in \eqref{eq:rco} is the numerator of Gower's
general coefficient of similarity~\cite{Gower1971}. Gower skips the attributes missing in either
object and divides by the number of compared attributes $|\Vobs{x}\cap\Vobs{y}|$. Hence $\RCO(x,y)$
is Gower's coefficient multiplied by $|\Vobs{x}\cap\Vobs{y}|/|A|$, the share of attributes that can
be compared. What is new here is the fixed denominator and the granules, dependency degree and
reduction built on it. The word ``overlap'' refers to the overlap of present-attribute sets;
$\RCO$ is not an overlap function in the sense of Bustince et al.~\cite{Bustince2010}, which
aggregates two membership degrees.

\begin{proposition}[Self-similarity and non-reflexivity]
\label{prop:self_similarity}
For any $x\in U$,
\begin{equation}
\RCO(x,x)=\frac{|\Vobs{x}|}{|A|}\leq 1,
\label{eq:reflexivity}
\end{equation}
with equality iff $x$ is complete. In particular $\RCO(x,x)<1$ whenever $x$ contains
missing values, so $\RCO$ is \emph{not} reflexive. Hence it is neither a fuzzy tolerance
(reflexive and symmetric) relation nor a similarity relation in the sense of
Radzikowska and Kerre~\cite{Radzikowska2002}, since both require reflexivity.
\end{proposition}

\begin{proof}
By Definition~\ref{def:rco}, $\Vobs{x}\cap \Vobs{x}=\Vobs{x}$, and
$\mu_a(x,x)=1$ from Definition~\ref{def:mu_a}. Hence
\[
\RCO(x,x)=\frac{1}{|A|}\sum_{a\in \Vobs{x}}1=\frac{|\Vobs{x}|}{|A|}\le 1.
\]
The last inequality is strict if and only if $\Vobs{x}\subsetneq A$,
i.e., $x$ has at least one missing attribute.
\end{proof}

\begin{proposition}[Symmetry]
\label{prop:symmetry}
For any $x,y\in U$,
\begin{equation}
\RCO(x,y)=\RCO(y,x).
\label{eq:symmetry}
\end{equation}
\end{proposition}

\begin{proof}
Using Definition~\ref{def:rco}, commutativity of intersection, and symmetry of
$\mu_a$ from Definition~\ref{def:mu_a}, we obtain
with $I=\Vobs{x}\cap \Vobs{y}$:
\[
\begin{aligned}
\RCO(x,y) &= \frac{1}{|A|}\sum_{a\in I}\mu_a(x,y)\\
          &= \frac{1}{|A|}\sum_{a\in I}\mu_a(y,x)=\RCO(y,x).
\end{aligned}
\]
\end{proof}

In the crisp case $\mu_a(x,y)=\mathbf{1}[f(x,a)=f(y,a)]$, $\RCO$ is transitive for the
\L{}ukasiewicz t-norm but neither $\min$- nor product-transitive, so it is not a fuzzy
$T$-equivalence either (Section~S1 of the Supplementary material).

\begin{theorem}[Subset monotonicity of $\RCO^B$]
\label{thm:monotonicity}
If $B\subseteq C\subseteq A$, then for all $x,y\in U$,
\begin{equation}
\RCO^B(x,y)\leq \RCO^C(x,y).
\label{eq:mono_rco}
\end{equation}
\end{theorem}

\begin{proof}
Since $B\subseteq C$, we have
$(\Vobs{x}\cap \Vobs{y})\cap B\subseteq(\Vobs{x}\cap \Vobs{y})\cap C$.
Let $I_B=(\Vobs{x}\cap \Vobs{y})\cap B$ and $I_C=(\Vobs{x}\cap \Vobs{y})\cap C$.
Since $I_B\subseteq I_C$ and $\mu_a(x,y)\ge 0$:
\[
\begin{aligned}
\RCO^B(x,y) &= \frac{1}{|A|}\sum_{a\in I_B}\mu_a(x,y)\\
            &\le \frac{1}{|A|}\sum_{a\in I_C}\mu_a(x,y)=\RCO^C(x,y).
\end{aligned}
\]
\end{proof}

\subsection{Granule and Approximation}
\label{subsec:granule}

\begin{definition}[$\delta$-Granule]
\label{def:granule}
For $B\subseteq A$ and threshold $\delta\in[0,1]$, define
\begin{equation}
[x]_{\delta}^{B}=\{y\in U:\RCO^B(x,y)\geq\delta\}.
\label{eq:delta_granule}
\end{equation}
When $B=A$ we drop the superscript and write $[x]_{\delta}=[x]_{\delta}^{A}$.
\end{definition}

\begin{definition}[Soft Consistency]
\label{def:consistency}
For each object $x\in U$ and $B\subseteq A$, define
\begin{equation}
\mathrm{consistency}_B(x)=
\frac{\sum_{y\in[x]_{\delta}^{B}}\RCO^B(x,y)\,\mathbf{1}[D(y)=D(x)]}
{\sum_{y\in[x]_{\delta}^{B}}\RCO^B(x,y)},
\label{eq:consistency}
\end{equation}
with the convention $\mathrm{consistency}_B(x)=0$ when $[x]_{\delta}^{B}=\varnothing$.
We abbreviate $\mathrm{consistency}(x)=\mathrm{consistency}_A(x)$.
\end{definition}

\begin{definition}[Soft Dependency Degree]
\label{def:gamma_soft}
For $B\subseteq A$, define
\begin{equation}
\gamma_{\mathrm{soft}}(B)=\frac{1}{|U|}\sum_{x\in U}\mathrm{consistency}_B(x).
\label{eq:gamma_soft}
\end{equation}
\end{definition}

The consistency \eqref{eq:consistency} is the lower approximation of the vaguely quantified rough
sets of Cornelis et al.~\cite{Cornelis2007VQRS} with the identity quantifier, applied to the
relation $\RCO^B$ cut at $\delta$. Cornelis and Jensen~\cite{Cornelis2008VQRSFS} used such
approximations for attribute selection. What COFRS adds is the relation $\RCO^B$ with the fixed
denominator $|A|$.

\begin{example}[Illustrative computation of $\RCO$ and $\gamma_{\mathrm{soft}}$]
\label{ex:rco_gamma}
Consider $U=\{x_1,x_2,x_3,x_4\}$, $A=\{a_1,a_2,a_3\}$ (all categorical, $|A|=3$),
and decision $D\in\{+,-\}$ (Table~\ref{tab:toy_is}).
Missing entries are marked ``--''.
Present-attribute sets:
$\Vobs{x_1}=\Vobs{x_2}=\{a_1,a_2,a_3\}$,
$\Vobs{x_3}=\{a_1,a_2\}$, $\Vobs{x_4}=\{a_1,a_3\}$.
Only $a\in \Vobs{x}\cap \Vobs{y}$ contribute to \eqref{eq:rco}
($\mu_a\in\{0,1\}$ on exact match).
E.g.\ $\Vobs{x_1}\cap \Vobs{x_3}=\{a_1,a_2\}$ gives $R_{13}=\tfrac{2}{3}$.
Table~\ref{tab:toy_rco} lists the full $\RCO$ matrix ($\RCO(x,y)=\RCO(y,x)$).

\begin{table}[pos=htbp]
\centering
\small
\caption{Fuzzy overlap similarity $\RCO$ for Example~\ref{ex:rco_gamma}.}
\label{tab:toy_rco}
\begin{tabular}{lcccc}
\toprule
$\RCO$ & $x_1$ & $x_2$ & $x_3$ & $x_4$ \\
\midrule
$x_1$ & $1$ & $\tfrac{2}{3}$ & $\tfrac{2}{3}$ & $\tfrac{1}{3}$ \\
$x_2$ & $\tfrac{2}{3}$ & $1$ & $\tfrac{1}{3}$ & $\tfrac{1}{3}$ \\
$x_3$ & $\tfrac{2}{3}$ & $\tfrac{1}{3}$ & $\tfrac{2}{3}$ & $0$ \\
$x_4$ & $\tfrac{1}{3}$ & $\tfrac{1}{3}$ & $0$ & $\tfrac{2}{3}$ \\
\bottomrule
\end{tabular}
\end{table}

Take $\delta=0.5$ and write $R_{ij}=\RCO(x_i,x_j)$.
For $x_1$, the $\delta$-granule is
$G_1=[x_1]_{0.5}=\{x_1,x_2,x_3\}$ (since $R_{14}=\tfrac{1}{3}<0.5$).
Only $x_1,x_3\in G_1$ share label $+$ with $x_1$.
The same-class weight is $1+\tfrac{2}{3}$ and the total granule weight
is $1+\tfrac{2}{3}+\tfrac{2}{3}$, hence
\begin{equation}
\mathrm{consistency}(x_1)=\tfrac{5}{7}\approx 0.714.
\nonumber
\end{equation}
For $x_2$, $G_2=\{x_1,x_2\}$; only $x_2$ has label $-$, so
$\mathrm{consistency}(x_2)=\tfrac{3}{5}$.
For $x_3$, $G_3=\{x_1,x_3\}$ gives $\mathrm{consistency}(x_3)=1$;
for $x_4$, $G_4=\{x_4\}$ gives $\mathrm{consistency}(x_4)=1$.
Therefore
\[
\gamma_{\mathrm{soft}}(A)=\tfrac{29}{35}\approx 0.829.
\]
Missing values reduce numerators (e.g.\ $\RCO(x_3,x_4)=0$) while $|A|$ stays fixed,
and mixed-class neighbours in $[x]_{\delta}$ lower soft consistency even when
$\RCO$ is moderately high (e.g.\ $x_1$ vs.\ $x_2$).
\end{example}

\begin{remark}[Self-membership and highly incomplete objects]
\label{rem:self_membership}
Because $\RCO$ is not reflexive in the classical sense
(Eq.~\eqref{eq:reflexivity}), self-similarity is
$\RCO^B(x,x)=|\VobsB{x}|/|A|$, so $x$ belongs to its own $\delta$-granule if and only if
$|\VobsB{x}|\geq\delta|A|$.
An object that misses more than a $(1-\delta)$ fraction of all attributes is therefore
excluded from its own granule. If its granule becomes empty, its consistency is $0$ by
convention, so $\gamma_{\mathrm{soft}}$ penalises such objects regardless of their labels.
This is a deliberate consequence of the conservative design: an object about which
almost nothing is observed should not certify dependency.
The thresholds chosen in our experiments are at most $0.30$ (Section~\ref{sec:experiments}), so with
$B=A$ only an object that misses more than $70\%$ of its attributes is affected.
The benchmarks contain sixteen such objects, in three of the eleven datasets.
The bound bites harder during backward elimination.
Since $\mu_a\leq 1$ implies $\RCO^B(x,y)\leq|B|/|A|$ for all $x,y$, every granule is empty and
$\gamma_{\mathrm{soft}}(B)=0$ whenever $|B|<\delta|A|$.
Candidate subsets smaller than $\lceil\delta|A|\rceil$ are thus rejected by the
$\varepsilon$-criterion of Algorithm~\ref{alg:cofrs_reduct}, which gives an implicit lower
bound on reduct size. It explains the monotone growth of $|B^*|$ with $\delta$ observed
in the $\delta$-sensitivity study (Figure~\ref{fig:diagnostics}b).
\end{remark}

We now define a graded lower approximation on $\RCO$ and show that $\gamma_{\mathrm{soft}}$ is the
average membership of the induced positive region. This places COFRS in the variable-precision
stream of rough sets~\cite{Ziarko1993,Yao2008}.

\begin{definition}[COFRS graded $\delta$-lower membership]
\label{def:cofrs_approx}
Let $D_k=\{y\in U:D(y)=k\}$ be a decision class and $[x]_\delta$ the $\delta$-granule
(Definition~\ref{def:granule}). The graded $\delta$-lower membership of $D_k$ at $x$ is
\begin{equation}
\underline{\RCO}_\delta(D_k)(x)
  = \frac{\sum_{y\in[x]_\delta}\RCO(x,y)\,\mathbf{1}[D(y)=k]}
          {\sum_{y\in[x]_\delta}\RCO(x,y)},
  \label{eq:cofrs_lower}
\end{equation}
i.e.\ the share of $\delta$-granule mass assigned to class $k$ (undefined, and by
convention $0$, when the granule is empty). The COFRS positive-region membership of $x$
is the lower membership of its own class,
$\mathrm{POS}_\delta(x)=\underline{\RCO}_\delta(D_{D(x)})(x)$.
For $B\subseteq A$, the $B$-restricted operator and $\mathrm{POS}_{\delta,B}(x)$ are
obtained by replacing $\RCO$ with $\RCO^B$ and $[x]_\delta$ with $[x]_\delta^{B}$
throughout.
\end{definition}

Because $\RCO$ is not reflexive, an object need not lie in its own granule, and the classical
sandwich between lower and upper approximations is lost. Over a crisp decision partition the
lower memberships of the classes sum to one, so a dual upper operator adds nothing, and we use
only the lower membership. For complete data, binary $\mu_a$ and $\delta\to1$, it reduces to the
Pawlak positive-region indicator (Section~S2 of the Supplementary material).

\begin{remark}[$\gamma_{\mathrm{soft}}$ as an averaged positive region]
\label{rem:gamma_pos}
By Definitions~\ref{def:consistency} and~\ref{def:cofrs_approx},
$\mathrm{POS}_{\delta,B}(x)=\mathrm{consistency}_B(x)$ for every $x$: the numerator of
\eqref{eq:cofrs_lower} restricted to the class $D(x)$ is the numerator of
\eqref{eq:consistency}, and the denominators coincide. Hence
\begin{equation}
\gamma_{\mathrm{soft}}(B)=\frac{1}{|U|}\sum_{x\in U}\mathrm{POS}_{\delta,B}(x),
\label{eq:gamma_pos}
\end{equation}
which is the COFRS analogue of the classical dependency degree
$\gamma_B(D)=|\mathrm{POS}_B(D)|/|U|$ \eqref{eq:dependency}.
\end{remark}

\begin{proposition}[Non-monotonicity of $\gamma_{\mathrm{soft}}$]
\label{prop:gamma_nonmono}
The map $B\mapsto\gamma_{\mathrm{soft}}(B)$ is \emph{not} monotone with respect to
attribute inclusion: there exist $B\subsetneq A$ with
$\gamma_{\mathrm{soft}}(B)>\gamma_{\mathrm{soft}}(A)$.
\end{proposition}

\begin{proof}
We give a closed-form counterexample.
Let $U=\{x_1,x_2,x_3\}$, $A=\{a,b\}$ (both categorical, $|A|=2$), all values present,
and decision labels $D(x_1)=D(x_2){=}{+}$, $D(x_3){=}{-}$:
\begin{center}
\small
\begin{tabular}{lcc}
\toprule
Object & $a$ & $b$ \\\midrule
$x_1$ & 1 & 0 \\
$x_2$ & 0 & 1 \\
$x_3$ & 1 & 1 \\\bottomrule
\end{tabular}
\end{center}
Since data are complete, $\Vobs{x_i}=A$ for all $i$ and
$\RCO^B(x_i,x_i)=|B|/|A|$ for any $B\subseteq A$.

\smallskip
\noindent\textbf{Full set $B=A$, $\delta=0.5$:}
$\RCO(x_1,x_2)=0$, $\RCO(x_1,x_3)=\tfrac{1}{2}$, $\RCO(x_2,x_3)=\tfrac{1}{2}$,
$\RCO(x_i,x_i)=1$.
Granules: $[x_1]_{0.5}=\{x_1,x_3\}$, $[x_2]_{0.5}=\{x_2,x_3\}$,
$[x_3]_{0.5}=\{x_1,x_2,x_3\}$.
\[
\mathrm{cons}(x_1)=\tfrac{2}{3},\qquad
\mathrm{cons}(x_2)=\tfrac{2}{3},\qquad
\mathrm{cons}(x_3)=\tfrac{1}{2}.
\]
Hence $\gamma_{\mathrm{soft}}(A)=\tfrac{1}{3}\!\left(\tfrac{2}{3}+\tfrac{2}{3}+\tfrac{1}{2}\right)=\tfrac{11}{18}\approx0.611$.

\smallskip
\noindent\textbf{Singleton $B=\{a\}$:}
$\RCO^{\{a\}}(x_i,x_i)=\tfrac{1}{2}$;
$\RCO^{\{a\}}(x_1,x_2)=0$, $\RCO^{\{a\}}(x_1,x_3)=\tfrac{1}{2}$, $\RCO^{\{a\}}(x_2,x_3)=0$.
Granules: $[x_1]_{0.5}=\{x_1,x_3\}$, $[x_2]_{0.5}=\{x_2\}$, $[x_3]_{0.5}=\{x_1,x_3\}$.
Now $x_2$ is isolated (no mixed-class neighbour), so $\mathrm{cons}(x_2)=1$, and
$\mathrm{cons}(x_1)=\mathrm{cons}(x_3)=\tfrac{1}{2}$.
Hence $\gamma_{\mathrm{soft}}(\{a\})=\tfrac{1}{3}\!\left(\tfrac{1}{2}+1+\tfrac{1}{2}\right)=\tfrac{2}{3}\approx0.667$.

\smallskip
Since $\{a\}\subsetneq A$ yet
$\gamma_{\mathrm{soft}}(\{a\})=\tfrac{2}{3}>\tfrac{11}{18}=\gamma_{\mathrm{soft}}(A)$,
monotonicity fails.
Here, adding attribute $b$ raises $\RCO$ for the pair $(x_2,x_3)$
from $0$ to $\tfrac{1}{2}$, admitting $x_3$ (label $-$) into $x_2$'s granule and
reducing $x_2$'s consistency from $1$ to $\tfrac{2}{3}$.
Theorem~\ref{thm:monotonicity} guarantees $\RCO^{\{a\}}\leq\RCO^A$ pointwise, but the
induced change in granule membership can strictly \emph{decrease} $\gamma_{\mathrm{soft}}$.
\end{proof}

\begin{remark}[Why monotonicity of $\RCO$ does not imply monotonicity of
$\gamma_{\mathrm{soft}}$]
\label{rem:nonmono_reconcile}
There is no contradiction between Theorem~\ref{thm:monotonicity}
($\RCO^B\leq\RCO^C$ for $B\subseteq C$) and Proposition~\ref{prop:gamma_nonmono}.
The similarity values grow monotonically with $B$. But
$\gamma_{\mathrm{soft}}$ depends on them only through the thresholded granule
$[x]_\delta$ and the class-purity ratio of \eqref{eq:cofrs_lower}.
Enlarging $B$ can pull a cross-class object across the $\delta$ cut into $[x]_\delta$, which lowers
$\mathrm{POS}_\delta(x)$ even as all $\RCO$ values increase. A positive region can also change non-monotonically in variable-precision and
tolerance-based rough sets~\cite{Ziarko1993,Kryszkiewicz1998}.
This is why Section~\ref{sec:algorithm} uses an $\varepsilon$-tolerant reduct instead of the
classical inclusion-based reduct.
\end{remark}

The next result gives two conditions under which monotonicity holds.

\begin{proposition}[Sufficient condition for monotonicity of $\gamma_{\mathrm{soft}}$]
\label{prop:gamma_mono_cond}
Let $B\subseteq C=B\cup\{b\}\subseteq A$ and write
$\Delta_b(x,y)=\mu_b(x,y)\,\mathbf{1}[b\in\Vobs{x}\cap\Vobs{y}]/|A|\ge 0$ for the similarity
mass contributed by the extra attribute $b$ (so $\RCO^C=\RCO^B+\Delta_b$ pointwise),
and let $w_x=\sum_{y\in[x]_\delta^{B}}\Delta_b(x,y)$ denote the total added granule mass at $x$,
the sum being taken over the $B$-granule (which equals the $C$-granule under
condition~(i) below).
Suppose that for every $x\in U$:
\begin{enumerate}[label=(\roman*)]
  \item \emph{(granule stability)} $[x]_\delta^{B}=[x]_\delta^{C}$, i.e.\ adding $b$ moves no
    object across the threshold $\delta$; and
  \item \emph{(purity alignment)} whenever $w_x>0$, the added mass is at least as class-pure
    as the current granule:
    \begin{equation}
    \frac{\sum_{y\in[x]_\delta^{B}}\Delta_b(x,y)\,\mathbf{1}[D(y)=D(x)]}{w_x}
    \;\ge\;\mathrm{consistency}_B(x).
    \label{eq:purity_align}
    \end{equation}
\end{enumerate}
Then $\gamma_{\mathrm{soft}}(C)\ge\gamma_{\mathrm{soft}}(B)$.
\end{proposition}

\begin{proof}
Fix $x$ and let $G=[x]_\delta^{B}=[x]_\delta^{C}$ by~(i). Put
\[
N=\!\!\sum_{y\in G}\!\RCO^B(x,y)\mathbf{1}[D(y){=}D(x)],\quad
W=\!\!\sum_{y\in G}\!\RCO^B(x,y),
\]
and $a=\sum_{y\in G}\Delta_b(x,y)\mathbf{1}[D(y){=}D(x)]$, $w=w_x=\sum_{y\in G}\Delta_b(x,y)$.
Since $\RCO^C=\RCO^B+\Delta_b$ and $G$ is unchanged,
$\mathrm{consistency}_C(x)=(N+a)/(W+w)$ while $\mathrm{consistency}_B(x)=N/W$.
If $w=0$ the two coincide. If $w>0$, condition~\eqref{eq:purity_align} reads $a/w\ge N/W$,
hence $aW\ge Nw$ and
\[
\mathrm{consistency}_C(x)-\mathrm{consistency}_B(x)
=\frac{aW-Nw}{W(W+w)}\;\ge\;0 .
\]
Thus $\mathrm{consistency}_C(x)\ge\mathrm{consistency}_B(x)$ for every $x$, and averaging over
$U$ (Definition~\ref{def:gamma_soft}) gives $\gamma_{\mathrm{soft}}(C)\ge\gamma_{\mathrm{soft}}(B)$.
\end{proof}

\begin{remark}[Interpretation and link to the search]
\label{rem:mono_cond}
The counterexample of Proposition~\ref{prop:gamma_nonmono} violates condition~(i): adding $b$
pulls a cross-class object into a granule. Under the conditions, adding $b$ cannot decrease
$\gamma_{\mathrm{soft}}$, so removing $b$ may decrease it. Backward elimination
(Algorithm~\ref{alg:cofrs_reduct}) therefore tests each removal directly against the
$\varepsilon$-tolerance.
\end{remark}

Moreover, $\gamma_{\mathrm{soft}}(\varnothing)=0$ and $\gamma_{\mathrm{soft}}(B)\in[0,1]$ for every
$B\subseteq A$ and every $\delta$ (Section~S1 of the Supplementary material).

\begin{remark}[Relation to min-based fuzzy rough sets]
\label{prop:frs_avg_relation}
If no value is missing, $\RCO(x,y)=\frac{1}{|A|}\sum_{a\in A}\mu_a(x,y)$, the average-aggregation
fuzzy rough similarity. Fuzzy-rough feature selection~\cite{Jensen2009} instead aggregates with
the $\min$ t-norm, and the classical approximations~\cite{DuboisPrade1990,Radzikowska2002} use
$\inf$--$\max$ compositions. COFRS differs in three ways. It averages over present attributes
with $|A|$ normalisation, it uses threshold granules instead of fuzzy lower approximations of
decision classes, and $\gamma_{\mathrm{soft}}$ measures weighted same-class consistency inside a
granule instead of the cardinality of a fuzzy positive region. With binary $\mu_a$, $\RCO$ is the
proportion of attributes on which two objects are both present and agree. It is a graded agreement
rate, not a wildcard tolerance relation. COFRS-min in Section~\ref{sec:experiments} replaces the
average by the minimum.
\end{remark}

\section{Algorithm}
\label{sec:algorithm}

The COFRS pipeline has three procedures. Algorithm~\ref{alg:compute_rco} builds the similarity
matrix $\RCO^B$ for an attribute subset $B$, Algorithm~\ref{alg:compute_gamma_soft} evaluates the soft
dependency degree $\gamma_{\mathrm{soft}}(B)$ from that matrix, and Algorithm~\ref{alg:cofrs_reduct}
searches for a compact approximate reduct by greedy backward elimination. Because the first two are
separate, the search can update the matrix for each candidate removal instead of recomputing it.
This gives the overall bound $O(|A|^2\cdot|U|^2)$ of Eq.~\eqref{eq:cost_search}.

\subsection{Computing the \texorpdfstring{$\RCO$}{R\_CO} Matrix}

Algorithm~\ref{alg:compute_rco} materialises the $|U|\times|U|$ symmetric similarity
matrix for a given subset $B\subseteq A$.
For each pair $(x,y)$ the algorithm first restricts attention to the attributes in $B$
that are observed in \emph{both} objects ($\VobsB{x}\cap\VobsB{y}$).
It then accumulates the per-attribute similarity $\mu_a$ over that intersection and divides by the
\emph{global} $|A|$, not by the intersection size.
The global denominator is the design choice that (a) penalises missingness
conservatively (Definition~\ref{def:rco}) and (b) preserves subset monotonicity
(Theorem~\ref{thm:monotonicity}).
If the intersection is empty, the sum is empty and the similarity is zero.

\begin{algorithm}[t]
\caption{Compute fuzzy overlap similarity matrix $\RCO^B$.}
\label{alg:compute_rco}
\begin{algorithmic}[1]
\REQUIRE $IS=(U,A,V,f)$, attribute subset $B\subseteq A$
\ENSURE Similarity matrix $R\in\mathbb{R}^{|U|\times |U|}$
\FORALL{$x\in U$}
  \STATE Compute $\VobsB{x}=\{a\in B: f(x,a)\neq *\}$
\ENDFOR
\FORALL{$(x,y)\in U\times U$}
  \STATE $\textit{common}\gets \VobsB{x}\cap \VobsB{y}$
  \IF{$\textit{common}=\varnothing$}
    \STATE $R[x][y]\gets 0$
  \ELSE
    \STATE $\textit{num}\gets \sum_{a\in \textit{common}} \mu_a(f(x,a),f(y,a))$
    \STATE $R[x][y]\gets \textit{num}/|A|$
  \ENDIF
\ENDFOR
\STATE \textbf{return} $R$
\end{algorithmic}
\end{algorithm}

The double loop over $U\times U$ dominates, and each pair sums at most $|B|$ attribute
similarities:
\begin{equation}
T_{\mathrm{sim}}(|U|,|B|)=O\big(|U|^2\cdot|B|\big).
\label{eq:cost_sim}
\end{equation}
In practice, the $\mu_a$ evaluations for continuous attributes use a pre-computed
per-attribute bandwidth $\sigma_a$ estimated on the training fold, so the constant
factor is small.

\subsection{Computing \texorpdfstring{$\gamma_{\mathrm{soft}}$}{gamma\_soft}}

Algorithm~\ref{alg:compute_gamma_soft} takes the pre-built similarity matrix $R$ as
input and computes $\gamma_{\mathrm{soft}}$ via Definitions~\ref{def:granule}
and~\ref{def:gamma_soft}.
For each object $x$ the algorithm thresholds the row $R[x,\cdot]$ at $\delta$ to form
the $\delta$-granule $[x]_\delta$, then computes the weighted fraction of same-class
similarity mass (soft consistency).
An empty granule is handled by setting consistency to zero and skipping to the next
object, as in Definition~\ref{def:consistency}.
The final score is the arithmetic mean of per-object consistency values.

\begin{algorithm}[t]
\caption{Compute soft dependency degree $\gamma_{\mathrm{soft}}$.}
\label{alg:compute_gamma_soft}
\begin{algorithmic}[1]
\REQUIRE Similarity matrix $R$, labels $D$, threshold $\delta$
\ENSURE $\gamma_{\mathrm{soft}}$ value
\STATE $\textit{total}\gets 0$
\FORALL{$x\in U$}
  \STATE $\textit{granule}\gets \{y\in U: R[x][y]\geq \delta\}$
  \IF{$\textit{granule}=\varnothing$}
    \STATE $\textit{consistency}\gets 0$
    \STATE \textbf{continue}
  \ENDIF
  \STATE $\textit{num}\gets \sum_{y\in \textit{granule}} R[x][y]\cdot \mathbf{1}[D(y)=D(x)]$
  \STATE $\textit{den}\gets \sum_{y\in \textit{granule}} R[x][y]$
  \STATE $\textit{consistency}\gets \textit{num}/\textit{den}$
  \STATE $\textit{total}\gets \textit{total}+\textit{consistency}$
\ENDFOR
\STATE \textbf{return} $\textit{total}/|U|$
\end{algorithmic}
\end{algorithm}

Given the similarity matrix, Algorithm~\ref{alg:compute_gamma_soft} iterates once over
all $|U|$ objects and, for each, scans the full row of $|U|$ similarities to determine
granule membership and accumulate weighted sums:
\begin{equation}
T_{\gamma}(|U|)=O\big(|U|^2\big).
\label{eq:cost_gamma}
\end{equation}
A higher $\delta$ gives smaller and purer granules. The value $\delta^*$ is selected by nested cross-validation
(Section~\ref{sec:experiments}).

\subsection{Finding Reduct}
\label{subsec:finding_reduct}

Reduct search in COFRS is formulated as an importance-guided backward elimination
problem under an $\varepsilon$-tolerance on dependency loss.

\begin{definition}[COFRS Attribute Importance]
\label{def:importance}
For $B\subseteq A$ and $a\in B$, the COFRS attribute importance of $a$
with respect to $B$ is
\begin{equation}
\sigma_B^{\mathrm{soft}}(a)
  =\gamma_{\mathrm{soft}}(B)-\gamma_{\mathrm{soft}}(B\setminus\{a\}).
\label{eq:importance}
\end{equation}
\end{definition}

\begin{remark}
Because $\gamma_{\mathrm{soft}}$ is not monotone
(Proposition~\ref{prop:gamma_nonmono}), $\sigma_B^{\mathrm{soft}}(a)$ may be
negative: removing $a$ can increase soft dependency, meaning the
attribute actively harms granule purity and should be eliminated.
\end{remark}

\begin{definition}[COFRS $\varepsilon$-Reduct]
\label{def:cofrs_reduct}
A subset $B^*\subseteq A$ is a COFRS $\varepsilon$-reduct if
$\gamma_{\mathrm{soft}}(B^*)\geq\gamma_{\mathrm{soft}}(A)-\varepsilon$
(dependency preserved within tolerance). It is \emph{locally compact} if, in addition,
no single removal $a\in B^*$ keeps both $\gamma_{\mathrm{soft}}(B^*\setminus\{a\})\geq
\gamma_{\mathrm{soft}}(A)-\varepsilon$ and $\gamma_{\mathrm{soft}}(B^*\setminus\{a\})>\tau$.
\end{definition}

\begin{remark}[Relation to classical reduct]
\label{rem:approx_reduct}
Classical reduct requires (i) $\gamma_B(D)=\gamma_A(D)$ and (ii) no proper subset
$B'\subset B$ satisfies (i).
Condition~(i) is ill-posed in COFRS because $\gamma_{\mathrm{soft}}$ is not monotone:
$\gamma_{\mathrm{soft}}(A)$ need not be the global maximum.
Definition~\ref{def:cofrs_reduct} replaces the equality constraint by an
$\varepsilon$-tolerance on the dependency drop, and minimality by local compactness with
respect to single removals. Algorithm~\ref{alg:cofrs_reduct} stops exactly when no single
removal is admissible, so its output is locally compact; it is not guaranteed to be minimal.
\end{remark}

Algorithm~\ref{alg:cofrs_reduct} finds a COFRS $\varepsilon$-reduct
(Definitions~\ref{def:importance}--\ref{def:cofrs_reduct}) via greedy backward elimination.
Starting from the full attribute set $A$, each round selects the attribute
$a^\star$ with the smallest importance $\sigma^{\mathrm{soft}}_{B^*}(a)$ of Definition~\ref{def:importance}
(line~7). Its deletion causes the \emph{smallest} drop in $\gamma_{\mathrm{soft}}$. It commits the removal only if the resulting dependency
satisfies two conditions (line~11):
\begin{enumerate}[label=(\roman*)]
  \item $\gamma_{\mathrm{try}}\geq\gamma_{A}-\varepsilon$, where $\gamma_{A}=
    \gamma_{\mathrm{soft}}(A)$ is the \emph{fixed} full-attribute reference computed once
    (line~3); and
  \item $\gamma_{\mathrm{try}}>\tau$, a degeneracy floor.
\end{enumerate}
Condition~(i) bounds the \emph{cumulative} drop from the full set by a single $\varepsilon$
(Remark~\ref{rem:tau_guard}). Condition~(ii) forbids subsets whose granules are so sparse that
$\gamma_{\mathrm{soft}}\approx 0$. We set $\tau=0.05$ throughout. Elimination stops when no
single removal satisfies both conditions, so the algorithm terminates in at most
$|A|-1$ removals.

The critical efficiency gain comes from the \emph{incremental update} on line~9.
Because $\RCO^B$ is additive in its attribute contributions
(Eq.~\eqref{eq:rco_b}), removing attribute $a$ from $B^*$ changes each matrix
entry by a known, closed-form amount:
\begin{equation}
\begin{aligned}
I_a(x,y)&:=\mathbf{1}\!\left[a\in \Vobs{x}\cap \Vobs{y}\right],\\
\Delta_a(x,y)&:=\frac{\mu_a(f(x,a),f(y,a))}{|A|}\,I_a(x,y),\\
R^{B_{\mathrm{try}}}[x][y]&=R^{B^*}[x][y]-\Delta_a(x,y).
\end{aligned}
\label{eq:incremental_update}
\end{equation}
The update costs
\begin{equation}
T_{\mathrm{upd}}(|U|)=O\big(|U|^2\big),
\label{eq:cost_upd}
\end{equation}
which avoids the full recomputation of cost $T_{\mathrm{sim}}(|U|,|B|)$ in
Eq.~\eqref{eq:cost_sim} that Algorithm~\ref{alg:compute_rco} would otherwise require per trial.
Setting $\varepsilon=0$ gives an exact criterion, and a larger $\varepsilon$ gives more compact
reducts. All experiments use $\varepsilon=0.001$.

\begin{remark}[Degeneracy guard and the fixed reference]
\label{rem:tau_guard}
If each step were compared with the running best value, drops of size at most $\varepsilon$
could compound and end at a subset whose granules are too sparse to certify any dependency.
The fixed reference $\gamma_{A}$ bounds the total drop by one $\varepsilon$, and the floor $\tau$
rejects every step into the region $\gamma_{\mathrm{soft}}\leq\tau$.
They change which attributes survive more than how many. The reduct size is bounded below by the
granule-emptiness floor $|B^*|\geq\lceil\delta|A|\rceil$ of Remark~\ref{rem:self_membership}.
In the experiments the search stops exactly at this floor in 50 of the 55 outer folds; in the other
five, all on Hypothyroid, the reduct has one attribute more.
The compactness of the reducts is therefore largely set by $\delta$ (Section~\ref{subsec:size_matched}).
\end{remark}

\begin{algorithm}[t]
\caption{Greedy backward elimination for a COFRS approximate reduct $B^*$.}
\label{alg:cofrs_reduct}
\begin{algorithmic}[1]
\REQUIRE $IS=(U,A,V,f,D)$, thresholds $\delta$, $\varepsilon$, degeneracy floor $\tau$
\ENSURE Approximate reduct $B^*\subseteq A$
\STATE $B^*\gets A$
\STATE $R^{B^*}\gets \text{Algorithm~\ref{alg:compute_rco}}(IS,B^*)$
\STATE $\gamma_{A}\gets \text{Algorithm~\ref{alg:compute_gamma_soft}}(R^{B^*},D,\delta)$
  \COMMENT{fixed full-set reference; \emph{not} updated during the search}
\STATE $\textit{improved}\gets \textbf{TRUE}$
\WHILE{$\textit{improved}$ \AND $|B^*|>1$}
  \STATE $\textit{improved}\gets \textbf{FALSE}$
  \STATE $a^\star\gets \arg\max_{a\in B^*}\gamma_{\mathrm{soft}}(B^*\setminus\{a\})$
    \COMMENT{smallest dependency drop; each candidate is evaluated with the update of Eq.~\eqref{eq:incremental_update}}
  \STATE $B_{\mathrm{try}}\gets B^*\setminus \{a^\star\}$
  \STATE $R^{B_{\mathrm{try}}}[x][y]\gets R^{B^*}[x][y] - \mu_{a^\star}(f(x,a^\star),f(y,a^\star))\cdot\mathbf{1}[a^\star\in \Vobs{x}\cap \Vobs{y}]/|A|$
    \COMMENT{$O(|U|^2)$ incremental update via Eq.~\eqref{eq:incremental_update}}
  \STATE $\gamma_{\mathrm{try}}\gets \text{Algorithm~\ref{alg:compute_gamma_soft}}(R^{B_{\mathrm{try}}},D,\delta)$
  \IF{$\gamma_{\mathrm{try}}\geq \gamma_{A}-\varepsilon$ \AND $\gamma_{\mathrm{try}}>\tau$}
    \STATE $B^*\gets B_{\mathrm{try}}$;\quad $R^{B^*}\gets R^{B_{\mathrm{try}}}$
    \STATE $\textit{improved}\gets \textbf{TRUE}$
  \ENDIF
\ENDWHILE
\STATE \textbf{return} $B^*$
\end{algorithmic}
\end{algorithm}

When several candidates give the same $\gamma_{\mathrm{soft}}(B^*\setminus\{a\})$, the
$\arg\max$ returns the attribute that comes first in the column order of the table.
The outer \textbf{while} loop executes at most $|A|$ rounds, one removal per round.
Each round scans at most $|A|$ candidates, and each candidate costs one incremental update and one
evaluation of $\gamma_{\mathrm{soft}}$.
The whole search therefore costs
\begin{equation}
T_{\mathrm{search}}(|U|,|A|)=O(|A|)\cdot O(|A|)\cdot\big(T_{\mathrm{upd}}+T_{\gamma}\big)
=O\big(|A|^2\cdot|U|^2\big),
\label{eq:cost_search}
\end{equation}
which is the bound stated in the abstract. The initial matrix $\RCO^{A}$ costs
$T_{\mathrm{sim}}(|U|,|A|)=O(|U|^2\cdot|A|)$ and is dominated by this term.

\section{Experimental Evaluation}
\label{sec:experiments}

The experiments test four claims about COFRS. The first is that it returns smaller reducts than the
tolerance and fuzzy-rough baselines. The second is that this advantage comes from the attributes
selected and not only from the size of the reduct. The third is that the reducts keep the downstream
accuracy, and the fourth is that the results do not hinge on the choice of $\delta$ or of the baseline
thresholds.

\subsection{Experimental Setup}

We evaluate on 11 UCI datasets with genuine missing values~\cite{UCI2023}
(Table~\ref{tab:datasets}; IR is the size of the largest class divided by that of the smallest).
Missing entries are those marked ``?''\ in the original files. On Pima, physiologically
impossible zeros in Glucose, BloodPressure, SkinThickness, Insulin and BMI are recoded as missing,
following common practice.
From Horse\_Colic we remove the hospital number, which identifies the case, and the four lesion
attributes recorded after surgery, which are not available when the decision is made.
Continuous attributes are min--max normalised once on the whole dataset. The bandwidths $\sigma_a$
and the ranges are then estimated on the training fold, except for FRFS, which estimates them on
the whole dataset. Reduct search uses the training fold only. The reduct does not depend on the
choice of scaling (Remark~\ref{rem:leakage}).
No imputation is performed at any stage of reduct computation; the classifiers use
median imputation fitted on the training fold.
The largest benchmark (Hypothyroid) contains 3163 objects. Every method builds an
$O(|U|^2)$ similarity matrix, so universes beyond about $10^4$ objects are not assessed here
(Section~\ref{sec:conclusion}).

\begin{remark}[Absence of preprocessing leakage]
\label{rem:leakage}
Fitting the min--max range on the training fold instead of the full sample would not change the
selected reducts. The Gaussian similarity $\exp(-d^2/\sigma_a^2)$ and the linear similarity
$\max(0,1-d/\mathrm{range}_a)$ do not change when $d$ and its bandwidth are rescaled by a common
factor. Only the $k$NN distances depend on this choice.
\end{remark}

\begin{table}[pos=htbp]
\centering
\footnotesize
\caption{Eleven UCI incomplete benchmarks (\#S, \#A, missing rate, IR).}
\label{tab:datasets}
\setlength{\tabcolsep}{3.5pt}
\begin{tabular}{lrrrrc}
\toprule
Dataset & \#S & \#A & Miss\% & IR & Source \\
\midrule
\multicolumn{6}{l}{\textit{High missing ($>5\%$) --- 6 datasets}} \\
Hepatitis       & 155 & 19 & 5.67  & 3.84 & UCI \\
Horse\_Colic & 299 & 22 & 24.35 & 4.05 & UCI \\
Pima            & 768 &  8 & 10.61 & 1.87 & UCI \\
Congressional   & 435 & 16 & 5.63  & 1.59 & UCI \\
Heart\_Hungarian& 294 & 13 & 20.46 & 1.77 & UCI \\
Hypothyroid     & 3163& 25 & 6.74  & 19.95 & UCI \\
\midrule
\multicolumn{6}{l}{\textit{Low missing ($\leq 5\%$) --- 5 datasets}} \\
Cleveland       & 303 & 13 & 0.15  & 1.18 & UCI \\
Credit          & 690 & 15 & 0.65  & 1.25 & UCI \\
Automobile      & 201 & 25 & 1.01  & 1.01 & UCI \\
Mammographic    & 961 &  5 & 3.37  & 1.16 & UCI \\
Cylinder\_Bands & 541 & 39 & 4.91  & 1.38 & UCI \\
\bottomrule
\end{tabular}
\end{table}

We compare COFRS with three crisp tolerance-based methods and an ablation on every benchmark,
plus a no-reduction control. Where a cited method does not cover
mixed incomplete data, we state the extension we used. All thresholds below are fixed, not tuned.
Each baseline tests one design choice. Tolerance-RS tests the wildcard semantics with a binary
relation, and Tol-RS$_\tau$ the same semantics with a threshold on continuous attributes.
Zhou-RS tests a binary similarity matrix restricted to object pairs that must be discerned.
COFRS-min tests the aggregation of the attribute similarities.
\textbf{Tolerance-RS}~\cite{Kryszkiewicz1998} implements the classical tolerance relation
with a binary similarity predicate in which missing values act as wildcards.
Two present values are related only if they are equal, which for continuous attributes means exact equality of the
normalised values. The reduct search is backward elimination that removes an attribute while
the dependency degree is unchanged.
\textbf{Tol-RS$_\tau$} extends this relation to continuous attributes with a range threshold.
Its relation is the neighbourhood-tolerance relation of Zhao and Qin~\cite{Zhao2014} with radius
$0.1\cdot\mathrm{range}(a)$; Zhao and Qin search with a conditional entropy, whereas here the crisp
dependency and the backward search of Tolerance-RS are used.
The range is estimated on the training fold.
\textbf{Zhou-RS} follows Algorithm~1 of Zhou and Bao~\cite{Zhou2023}.
On a categorical attribute two objects are similar if they are equal or if either value is missing.
Zhou and Bao treat categorical data only; on a continuous attribute we call two objects similar if
their values differ by at most $0.5$ standard deviations or if either is missing.
The binary similarity matrix has one row for each pair of objects from different classes of which
at least one is consistent, and one column per attribute.
The algorithm keeps the core attributes, then repeatedly deletes the attribute that leaves the
most pairs similar, together with the rows it fails to discern, until no similar pair is left.
As in the original, the result is not repaired when it does not preserve the consistent part;
this happens in 10 of the 55 folds. On Cleveland the result is empty in all 5 folds,
and we then use all attributes (counted as $|C|$ in the tables).
\textbf{COFRS-min} replaces the overlap average of COFRS by the $\min$ operator and uses the same
$\gamma_{\mathrm{soft}}$ search at a fixed $\delta=0.20$, a value from the grid used for COFRS.
It is \emph{not} the Jensen--Shen algorithm~\cite{Jensen2004}, which uses the fuzzy positive region.
That algorithm and the neighbourhood method of Wu et al.~\cite{Wu2024} are reimplemented separately
(Section~\ref{subsec:faithful_baselines}). The neighbourhood method of Li et al.~\cite{Li2023} was not
reimplemented. The \textbf{Full Set} baseline performs no attribute reduction.

The granule threshold $\delta$ is selected by nested cross-validation on each training fold:
$\delta^* = \arg\max_\delta \text{G-mean}$ over candidates
$\delta \in \{0.05, 0.10, 0.15, 0.20, 0.25, 0.30\}$, with ties resolved towards the smaller $\delta$.
Table~S1 of the Supplementary material reports the selected $\delta^*$ per fold.

\begin{remark}[$\delta$ selection and indirect target leakage]
\label{rem:delta_leakage}
Selecting $\delta^*$ on validation folds inside each training fold does not expose test labels.
Because G-mean is a downstream metric, $\delta$ is tuned for classification rather than purely for
reduction quality, and the reduct size follows $\delta$ (Remark~\ref{rem:tau_guard}).
The comparison is not symmetric: COFRS selects $\delta$ per fold, whereas the thresholds of the
baselines are fixed. Figure~\ref{fig:diagnostics}b shows the sensitivity to $\delta$.
\end{remark}

During greedy backward elimination (Algorithm~\ref{alg:cofrs_reduct}), we set
$\varepsilon = 0.001$ across all datasets without further tuning.
Downstream evaluation uses $k$NN ($k{=}5$) and RBF-SVM (default $C{=}1$), averaged over
stratified 5-fold CV (seed 42). Standard deviations should be read with care on
the smallest benchmarks (Hepatitis, $n{=}155$; Automobile, $n{=}201$).
All methods were implemented in Python 3.9.6 with NumPy 2.0.2, SciPy 1.13.1 and
scikit-learn 1.6.1, on a machine with an Apple M4 Pro processor (14 cores), 48~GB of RAM
and macOS 26.6.2.

\subsection{Main Results}
\label{subsec:main_results}

COFRS gives small reducts, and its gain is compactness, not accuracy.
With a $k$NN classifier its accuracy, F1 and G-mean do not differ significantly from those of the
baselines, whereas with an RBF-SVM it has the lowest average ranks
(Section~\ref{subsec:statistical_analysis}).
The reduct size is set largely by $\delta$: the search stops at the granule floor
$\lceil\delta|A|\rceil$ in 50 of the 55 outer folds (Section~\ref{subsec:finding_reduct}).
At matched cardinality, COFRS selects better attributes than Tol-RS and random subsets
(Section~\ref{subsec:size_matched}).

\begin{figure}[pos=htbp]
\centering
\includegraphics[width=0.92\linewidth]{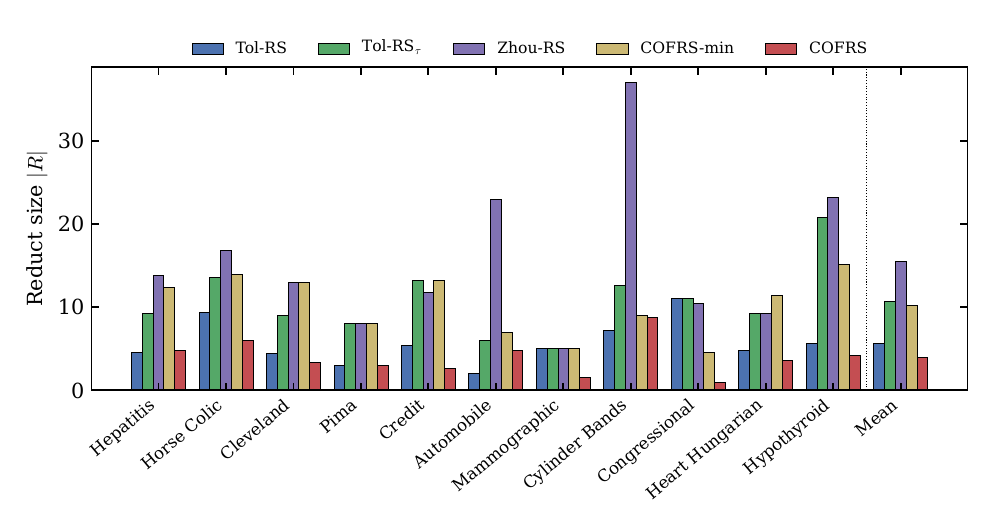}
\caption{Reduct size $|R|$ per dataset (mean over 5 folds) for the five reduction methods,
with the mean over the 11 datasets at the bottom. COFRS is shown in red.}
\label{fig:reduct_per_dataset}
\end{figure}

\paragraph{Attribute reduction quality.}
Table~\ref{tab:reduct_size} reports the reduct size (mean$\pm$std over 5-fold CV), and
Figure~\ref{fig:reduct_per_dataset} plots the same means per dataset.

\begin{table}[pos=htbp]
\centering
\caption{Reduct size $|B|$ (mean$\pm$std, 5-fold CV). \textbf{Bold}: smallest reduct among reduction methods.}
\label{tab:reduct_size}
\footnotesize
\begin{tabular}{lrrrrrr}
\toprule
Dataset & Tol-RS & \fRIC & Zhou-RS & \fRFS & Full & COFRS \\
\midrule
Hepatitis        & \textbf{4.6$\pm$0.55} & 9.2$\pm$0.84 & 13.8$\pm$1.10 & 12.4$\pm$0.55 & 19 & 4.8$\pm$2.17 \\
Horse\_Colic & 9.4$\pm$0.89 & 13.6$\pm$0.55 & 16.8$\pm$0.84 & 14.0$\pm$1.22 & 22 & \textbf{6.0$\pm$1.22} \\
Cleveland        & 4.4$\pm$0.89 & 9.0$\pm$0.71 & 13.0$\pm$0.00 & 13.0$\pm$0.00 & 13 & \textbf{3.4$\pm$0.55} \\
Pima             & \textbf{3.0$\pm$0.00} & 8.0$\pm$0.00 & 8.0$\pm$0.00 & 8.0$\pm$0.00 & 8 & \textbf{3.0$\pm$0.00} \\
Credit           & 5.4$\pm$0.89 & 13.2$\pm$0.84 & 11.8$\pm$1.30 & 13.2$\pm$0.45 & 15 & \textbf{2.6$\pm$0.89} \\
Automobile       & \textbf{2.0$\pm$0.00} & 6.0$\pm$0.00 & 23.0$\pm$0.00 & 7.0$\pm$0.00 & 25 & 4.8$\pm$2.05 \\
Mammographic     & 5.0$\pm$0.00 & 5.0$\pm$0.00 & 5.0$\pm$0.00 & 5.0$\pm$0.00 & 5 & \textbf{1.6$\pm$0.55} \\
Cylinder\_Bands  & \textbf{7.2$\pm$0.84} & 12.6$\pm$0.89 & 37.0$\pm$0.00 & 9.0$\pm$6.16 & 39 & 8.8$\pm$2.68 \\
Congressional    & 11.0$\pm$1.00 & 11.0$\pm$1.00 & 10.4$\pm$2.07 & 4.6$\pm$4.16 & 16 & \textbf{1.0$\pm$0.00} \\
Heart\_Hungarian & 4.8$\pm$0.45 & 9.2$\pm$0.45 & 9.2$\pm$2.59 & 11.4$\pm$0.55 & 13 & \textbf{3.6$\pm$0.55} \\
Hypothyroid      & 5.6$\pm$0.55 & 20.8$\pm$0.45 & 23.2$\pm$0.45 & 15.2$\pm$0.45 & 25 & \textbf{4.2$\pm$0.84} \\
\tabaggseparator
\tabaggrow & 5.7 & 10.7 & 15.6 & 10.3 & 18.2 & \textbf{4.0} \\
\bottomrule
\end{tabular}
\end{table}

\begin{figure}[pos=htbp]
\centering
\begin{subfigure}[t]{\linewidth}
\centering
\includegraphics[width=0.88\linewidth]{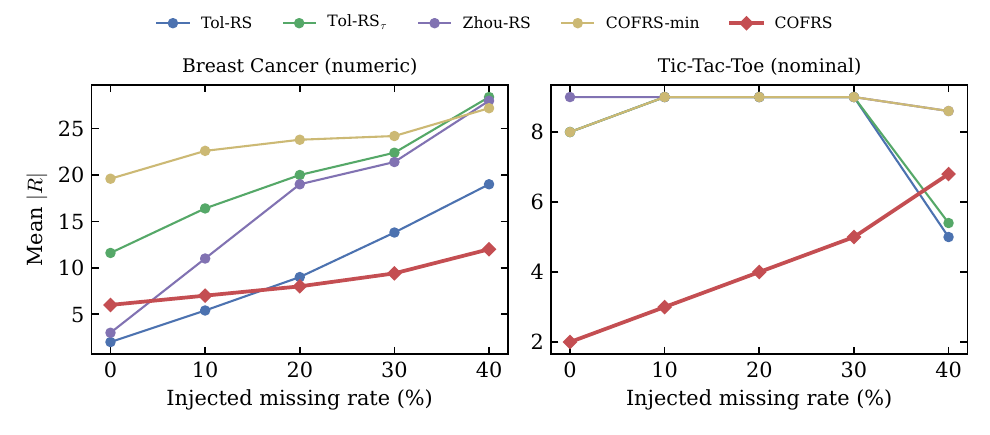}
\caption{Robustness to injected missingness.}
\label{fig:robustness}
\end{subfigure}
\par\medskip
\begin{subfigure}[t]{0.70\linewidth}
\centering
\includegraphics[width=0.88\linewidth]{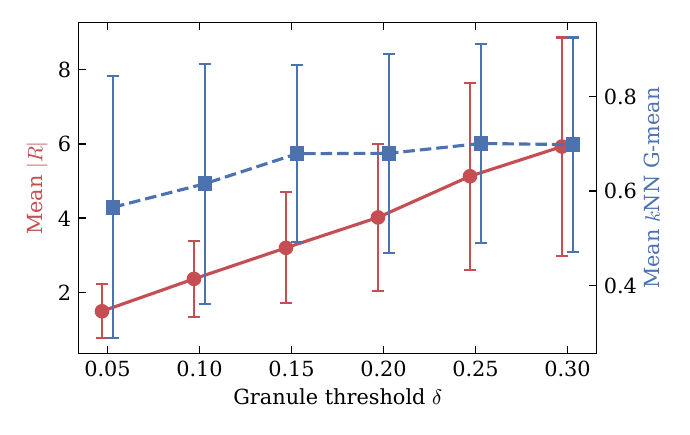}
\caption{Sensitivity to the granule threshold $\delta$.}
\label{fig:delta_sensitivity}
\end{subfigure}
\caption{Diagnostic analyses of COFRS with $k$NN. \textbf{(a)}~Mean $|R|$ under injected MCAR missingness.
\textbf{(b)}~Mean $|R|$ and $k$NN G-mean against $\delta$ (mean$\pm$1\,s.d.\ over 11 datasets).}
\label{fig:diagnostics}
\end{figure}

COFRS yields the smallest reduct on 8 of 11 datasets (one tie) and has the lowest mean
($4.0$ attributes, against $5.7$ for Tol-RS and $10.3$--$15.6$ for the other three).
Tol-RS is smaller on the other three datasets (Figure~\ref{fig:reduct_per_dataset}).
The size reduction is significant against Tol-RS$_\tau$, Zhou-RS and COFRS-min (Holm-adjusted
Wilcoxon $p<0.01$). It is not significant against Tol-RS ($p=0.14$), on which COFRS is smaller on
7 datasets, equal on 1 and larger on 3.
Against Tol-RS the clearer distinction is reduct quality rather than size
(Section~\ref{subsec:size_matched}).
Tol-RS$_\tau$ and Zhou-RS return large reducts because their wildcard treatment of missing values
inflates similarity, so more attributes are needed to separate the classes.
Per-fold tuning of $\delta$ is not what makes the reducts small.
At the fixed $\delta=0.20$ used for COFRS-min, COFRS returns $4.0$ attributes on average
(Figure~\ref{fig:diagnostics}b), whereas COFRS-min returns $10.3$.
The two differ only in the aggregation of the attribute similarities.

\paragraph{Sensitivity to the baseline thresholds.}
\label{subsec:baseline_sensitivity}
The thresholds of the baselines are fixed, so we also varied each of them over a grid
(Table~S3 of the Supplementary material).
The smallest mean reduct size among the 16 settings is $8.95$ attributes, more than twice that of
COFRS, and the mean $k$NN G-mean stays between $0.67$ and $0.72$ for all settings.
A stricter threshold therefore neither closes the size gap nor improves the downstream score.

\paragraph{Cross-fold stability.}
\label{subsec:reduct_stability}
Table~S5 of the Supplementary material gives the mean pairwise Jaccard similarity of the reducts
selected on the five outer folds.
COFRS is the least stable method on average ($0.59$, against $0.73$--$0.88$ for the
baselines), although it is perfectly stable on Congressional.
With few attributes, replacing one of them on another fold changes a large share of the set,
whereas Zhou-RS is the most stable largely because it keeps most attributes.

\begin{figure}[pos=htbp]
\centering
\includegraphics[width=0.92\linewidth]{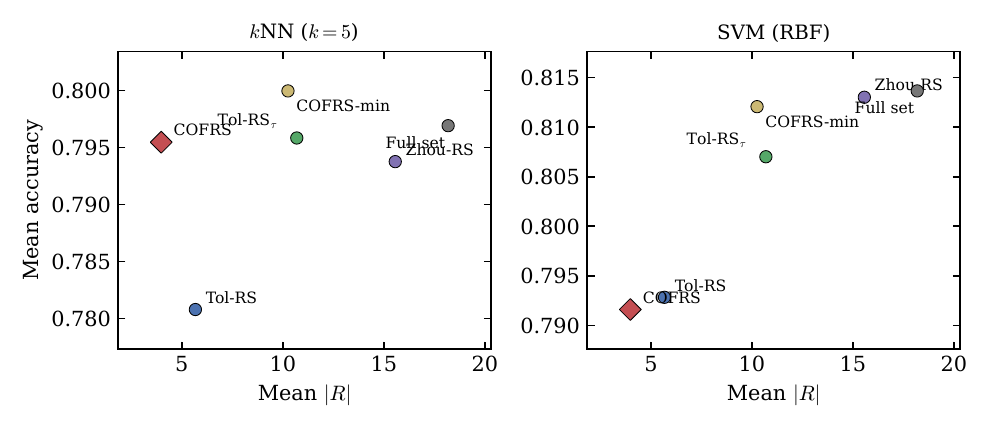}
\caption{Mean reduct size $|R|$ against mean accuracy over the 11 datasets, for $k$NN
($k{=}5$, left) and SVM (RBF, right). Each marker is one method; COFRS is the red diamond.}
\label{fig:tradeoff}
\end{figure}

\paragraph{Classification performance.}
Tables~\ref{tab:accuracy} and~\ref{tab:gmean} report accuracy and G-mean, and Table~S2 of the Supplementary material reports F1
(mean$\pm$std, 5-fold CV; $k$NN $k{=}5$, SVM-RBF).
Bold/underline mark best/second-best (ties in bold); W/T/L uses $\varepsilon{=}0.01$.
In the compactness--accuracy plane of Figure~\ref{fig:tradeoff}, COFRS has the smallest mean
reduct. Its mean $k$NN accuracy ($0.795$) is within $0.01$ of that of every other method except
Tol-RS, which it exceeds by $0.015$.
Its mean SVM accuracy ($0.792$) is about $0.02$ below those of Tol-RS$_\tau$, Zhou-RS, COFRS-min
and the full set.

\begin{table}[pos=htbp]
\centering
\caption{Classification accuracy on reduct-induced features (mean$\pm$std, 5-fold CV).}
\label{tab:accuracy}
\footnotesize
\setlength{\tabcolsep}{3pt}
\resizebox{\linewidth}{!}{
\begin{tabular}{@{}lcccccc|cccccc@{}}
\toprule
& \multicolumn{6}{c|}{KNN ($k=5$)} & \multicolumn{6}{c}{SVM (RBF)} \\
\cmidrule(lr){2-7}\cmidrule(lr){8-13}
Dataset & Tol & \fRIC & Zhou & \fRFS & COFRS & Full & Tol & \fRIC & Zhou & \fRFS & COFRS & Full \\
\midrule
Hepatitis & \underline{0.81$\pm$0.05} & 0.79$\pm$0.09 & \underline{0.81$\pm$0.07} & \underline{0.81$\pm$0.02} & 0.77$\pm$0.07 & \textbf{0.83$\pm$0.05} & \underline{0.83$\pm$0.01} & 0.79$\pm$0.08 & \underline{0.83$\pm$0.05} & 0.82$\pm$0.02 & 0.79$\pm$0.09 & \textbf{0.85$\pm$0.04} \\
Horse\_Colic & 0.60$\pm$0.05 & 0.57$\pm$0.05 & 0.59$\pm$0.04 & \textbf{0.63$\pm$0.06} & \underline{0.62$\pm$0.06} & 0.60$\pm$0.04 & 0.64$\pm$0.04 & 0.62$\pm$0.02 & 0.64$\pm$0.07 & \textbf{0.66$\pm$0.05} & \underline{0.65$\pm$0.06} & 0.63$\pm$0.06 \\
Cleveland & 0.77$\pm$0.05 & \textbf{0.82$\pm$0.04} & \textbf{0.82$\pm$0.03} & \textbf{0.82$\pm$0.03} & \underline{0.80$\pm$0.03} & \textbf{0.82$\pm$0.03} & 0.76$\pm$0.07 & \textbf{0.82$\pm$0.06} & \textbf{0.82$\pm$0.04} & \textbf{0.82$\pm$0.04} & \underline{0.79$\pm$0.07} & \textbf{0.82$\pm$0.04} \\
Pima & 0.65$\pm$0.04 & \underline{0.73$\pm$0.03} & \underline{0.73$\pm$0.03} & \underline{0.73$\pm$0.03} & \textbf{0.74$\pm$0.04} & \underline{0.73$\pm$0.03} & 0.70$\pm$0.02 & \textbf{0.77$\pm$0.02} & \textbf{0.77$\pm$0.02} & \textbf{0.77$\pm$0.02} & \underline{0.76$\pm$0.03} & \textbf{0.77$\pm$0.02} \\
Credit & 0.72$\pm$0.05 & 0.80$\pm$0.04 & \underline{0.82$\pm$0.03} & 0.80$\pm$0.04 & \textbf{0.84$\pm$0.04} & 0.80$\pm$0.04 & \underline{0.73$\pm$0.05} & \textbf{0.86$\pm$0.04} & \textbf{0.86$\pm$0.04} & \textbf{0.86$\pm$0.04} & \textbf{0.86$\pm$0.04} & \textbf{0.86$\pm$0.04} \\
Automobile & \textbf{0.91$\pm$0.06} & \underline{0.85$\pm$0.06} & 0.83$\pm$0.03 & 0.84$\pm$0.04 & 0.84$\pm$0.03 & 0.83$\pm$0.02 & \textbf{0.90$\pm$0.05} & 0.83$\pm$0.05 & \underline{0.86$\pm$0.05} & 0.85$\pm$0.04 & 0.76$\pm$0.18 & \underline{0.86$\pm$0.05} \\
Mammographic & \underline{0.78$\pm$0.01} & \underline{0.78$\pm$0.01} & \underline{0.78$\pm$0.01} & \underline{0.78$\pm$0.01} & \textbf{0.79$\pm$0.03} & \underline{0.78$\pm$0.01} & \textbf{0.80$\pm$0.02} & \textbf{0.80$\pm$0.02} & \textbf{0.80$\pm$0.02} & \textbf{0.80$\pm$0.02} & \underline{0.78$\pm$0.02} & \textbf{0.80$\pm$0.02} \\
Cylinder Bands & \underline{0.66$\pm$0.04} & \textbf{0.71$\pm$0.03} & 0.65$\pm$0.03 & \underline{0.66$\pm$0.03} & 0.65$\pm$0.03 & 0.65$\pm$0.03 & 0.63$\pm$0.03 & \textbf{0.65$\pm$0.01} & \underline{0.64$\pm$0.03} & \underline{0.64$\pm$0.04} & \underline{0.64$\pm$0.03} & \underline{0.64$\pm$0.03} \\
Congressional & 0.93$\pm$0.04 & 0.93$\pm$0.04 & 0.93$\pm$0.03 & \underline{0.95$\pm$0.04} & \textbf{0.96$\pm$0.03} & 0.93$\pm$0.03 & \textbf{0.96$\pm$0.03} & \textbf{0.96$\pm$0.03} & \textbf{0.96$\pm$0.04} & \textbf{0.96$\pm$0.03} & \textbf{0.96$\pm$0.03} & \textbf{0.96$\pm$0.03} \\
Heart Hungarian & 0.78$\pm$0.06 & \underline{0.80$\pm$0.03} & \underline{0.80$\pm$0.03} & \textbf{0.82$\pm$0.03} & 0.78$\pm$0.05 & \textbf{0.82$\pm$0.02} & 0.81$\pm$0.05 & \underline{0.82$\pm$0.04} & \textbf{0.83$\pm$0.06} & 0.81$\pm$0.04 & 0.78$\pm$0.07 & \underline{0.82$\pm$0.04} \\
Hypothyroid & \textbf{0.97$\pm$0.01} & \textbf{0.97$\pm$0.01} & \textbf{0.97$\pm$0.01} & \textbf{0.97$\pm$0.01} & \textbf{0.97$\pm$0.01} & \textbf{0.97$\pm$0.01} & \textbf{0.97$\pm$0.01} & \textbf{0.97$\pm$0.00} & \textbf{0.97$\pm$0.00} & \textbf{0.97$\pm$0.00} & \underline{0.96$\pm$0.01} & \underline{0.96$\pm$0.00} \\
\tabaggseparator
\tabaggrow & 0.78$\pm$0.11 & 0.80$\pm$0.10 & 0.79$\pm$0.10 & 0.80$\pm$0.10 & 0.80$\pm$0.10 & 0.80$\pm$0.10 & 0.79$\pm$0.11 & 0.81$\pm$0.10 & 0.81$\pm$0.10 & 0.81$\pm$0.10 & 0.79$\pm$0.10 & 0.81$\pm$0.10 \\
\bottomrule
\end{tabular}
}
\par\smallskip
{\footnotesize
\setlength{\tabcolsep}{5pt}
\begin{tabular}{@{}lccccc@{}}
\toprule
COFRS W/T/L vs. & Tol & \fRIC & Zhou & \fRFS & Full \\
\midrule
KNN ($k{=}5$) & 5/4/2 & 3/3/5 & 3/5/3 & 1/7/3 & 3/5/3 \\
SVM (RBF) & 3/4/4 & 1/5/5 & 0/6/5 & 0/5/6 & 1/5/5 \\
\bottomrule
\end{tabular}}
\end{table}

\begin{table}[pos=htbp]
\centering
\caption{G-mean on reduct-induced features (mean$\pm$std, 5-fold CV).}
\label{tab:gmean}
\footnotesize
\setlength{\tabcolsep}{3pt}
\resizebox{\linewidth}{!}{
\begin{tabular}{@{}lcccccc|cccccc@{}}
\toprule
& \multicolumn{6}{c|}{KNN ($k=5$)} & \multicolumn{6}{c}{SVM (RBF)} \\
\cmidrule(lr){2-7}\cmidrule(lr){8-13}
Dataset & Tol & \fRIC & Zhou & \fRFS & COFRS & Full & Tol & \fRIC & Zhou & \fRFS & COFRS & Full \\
\midrule
Hepatitis & 0.58$\pm$0.11 & 0.60$\pm$0.15 & 0.58$\pm$0.13 & \underline{0.62$\pm$0.15} & 0.53$\pm$0.30 & \textbf{0.67$\pm$0.10} & 0.51$\pm$0.31 & 0.55$\pm$0.12 & \underline{0.64$\pm$0.07} & 0.58$\pm$0.13 & 0.53$\pm$0.31 & \textbf{0.70$\pm$0.03} \\
Horse\_Colic\textsuperscript{a} & \underline{0.28$\pm$0.16} & \underline{0.28$\pm$0.17} & 0.00$\pm$0.00 & \textbf{0.29$\pm$0.17} & 0.21$\pm$0.19 & 0.00$\pm$0.00 & \underline{0.00$\pm$0.00} & \underline{0.00$\pm$0.00} & \underline{0.00$\pm$0.00} & \textbf{0.07$\pm$0.16} & \underline{0.00$\pm$0.00} & \underline{0.00$\pm$0.00} \\
Cleveland & 0.76$\pm$0.05 & \textbf{0.81$\pm$0.04} & \textbf{0.81$\pm$0.02} & \textbf{0.81$\pm$0.02} & \underline{0.79$\pm$0.03} & \textbf{0.81$\pm$0.02} & 0.75$\pm$0.08 & \textbf{0.82$\pm$0.06} & \underline{0.81$\pm$0.04} & \underline{0.81$\pm$0.04} & 0.77$\pm$0.07 & \underline{0.81$\pm$0.04} \\
Pima & \underline{0.59$\pm$0.05} & \textbf{0.69$\pm$0.03} & \textbf{0.69$\pm$0.03} & \textbf{0.69$\pm$0.03} & \textbf{0.69$\pm$0.05} & \textbf{0.69$\pm$0.03} & 0.58$\pm$0.03 & \textbf{0.69$\pm$0.03} & \textbf{0.69$\pm$0.03} & \textbf{0.69$\pm$0.03} & \underline{0.68$\pm$0.05} & \textbf{0.69$\pm$0.03} \\
Credit & 0.70$\pm$0.06 & 0.79$\pm$0.05 & \underline{0.81$\pm$0.04} & 0.79$\pm$0.05 & \textbf{0.83$\pm$0.04} & 0.79$\pm$0.05 & \underline{0.68$\pm$0.08} & \textbf{0.86$\pm$0.04} & \textbf{0.86$\pm$0.04} & \textbf{0.86$\pm$0.04} & \textbf{0.86$\pm$0.04} & \textbf{0.86$\pm$0.04} \\
Automobile & \textbf{0.90$\pm$0.06} & \underline{0.84$\pm$0.06} & 0.83$\pm$0.03 & \underline{0.84$\pm$0.04} & 0.83$\pm$0.03 & 0.83$\pm$0.02 & \textbf{0.90$\pm$0.06} & 0.83$\pm$0.04 & \underline{0.85$\pm$0.05} & 0.84$\pm$0.04 & 0.74$\pm$0.21 & \underline{0.85$\pm$0.05} \\
Mammographic & \textbf{0.78$\pm$0.01} & \textbf{0.78$\pm$0.01} & \textbf{0.78$\pm$0.01} & \textbf{0.78$\pm$0.01} & \textbf{0.78$\pm$0.04} & \textbf{0.78$\pm$0.01} & \textbf{0.79$\pm$0.02} & \textbf{0.79$\pm$0.02} & \textbf{0.79$\pm$0.02} & \textbf{0.79$\pm$0.02} & \underline{0.77$\pm$0.03} & \textbf{0.79$\pm$0.02} \\
Cylinder Bands & \underline{0.64$\pm$0.04} & \textbf{0.69$\pm$0.03} & 0.62$\pm$0.05 & \underline{0.64$\pm$0.03} & 0.62$\pm$0.04 & 0.62$\pm$0.05 & 0.36$\pm$0.08 & \underline{0.44$\pm$0.04} & 0.42$\pm$0.06 & \textbf{0.47$\pm$0.10} & 0.42$\pm$0.06 & 0.42$\pm$0.06 \\
Congressional & 0.92$\pm$0.04 & 0.92$\pm$0.04 & 0.93$\pm$0.04 & \underline{0.95$\pm$0.05} & \textbf{0.96$\pm$0.03} & 0.93$\pm$0.03 & \textbf{0.96$\pm$0.03} & \textbf{0.96$\pm$0.03} & \textbf{0.96$\pm$0.04} & \underline{0.95$\pm$0.04} & \textbf{0.96$\pm$0.03} & \textbf{0.96$\pm$0.03} \\
Heart Hungarian & 0.73$\pm$0.07 & 0.74$\pm$0.04 & 0.75$\pm$0.05 & \underline{0.78$\pm$0.06} & 0.74$\pm$0.06 & \textbf{0.79$\pm$0.04} & \underline{0.77$\pm$0.05} & \underline{0.77$\pm$0.05} & \textbf{0.79$\pm$0.06} & 0.76$\pm$0.05 & 0.72$\pm$0.07 & 0.76$\pm$0.05 \\
Hypothyroid & \underline{0.79$\pm$0.09} & 0.69$\pm$0.11 & 0.69$\pm$0.11 & 0.68$\pm$0.12 & \textbf{0.81$\pm$0.09} & 0.69$\pm$0.11 & \underline{0.58$\pm$0.14} & \textbf{0.59$\pm$0.09} & 0.54$\pm$0.05 & 0.56$\pm$0.07 & 0.40$\pm$0.30 & 0.53$\pm$0.06 \\
\tabaggseparator
\tabaggrow & 0.70$\pm$0.17 & 0.71$\pm$0.16 & 0.68$\pm$0.24 & 0.71$\pm$0.16 & 0.71$\pm$0.19 & 0.69$\pm$0.23 & 0.63$\pm$0.26 & 0.66$\pm$0.25 & 0.67$\pm$0.26 & 0.67$\pm$0.24 & 0.62$\pm$0.26 & 0.67$\pm$0.26 \\
\bottomrule
\end{tabular}
}
\par\smallskip
{\footnotesize
\setlength{\tabcolsep}{5pt}
\begin{tabular}{@{}lccccc@{}}
\toprule
COFRS W/T/L vs. & Tol & \fRIC & Zhou & \fRFS & Full \\
\midrule
KNN ($k{=}5$) & 5/2/4 & 3/3/5 & 4/4/3 & 3/3/5 & 4/4/3 \\
SVM (RBF) & 5/2/4 & 0/4/7 & 0/5/6 & 0/3/8 & 0/5/6 \\
\bottomrule
\end{tabular}}
\par\smallskip
{\footnotesize\textsuperscript{a}G-mean${}=0$ indicates that SVM predicts a single class for all test instances; Horse\_Colic has three classes (the smallest holds 15\% of the objects) and 24\% missing values.}
\end{table}

\paragraph{Robustness to injected missingness and sensitivity to $\delta$.}
We injected missing values completely at random, at rates from $0$ to $40\%$, into the complete
datasets Breast\_Cancer (numeric attributes) and Tic-Tac-Toe (nominal attributes) and recomputed the
reducts (Figure~\ref{fig:diagnostics}a).
On Breast\_Cancer the COFRS reduct grows from $6.0$ to $12.0$ attributes.
The four baseline reducts grow faster and end between $19.0$ and $28.4$ attributes at $40\%$,
including those of Tol-RS and Zhou-RS, which start below COFRS.
On Tic-Tac-Toe COFRS grows from $2.0$ to $6.8$ attributes, while the baselines stay between
$8$ and $9$ up to $30\%$ and fall to $5.0$--$8.6$ at $40\%$.
The conservative penalty therefore limits the growth of the reduct on numeric data and does not
do so on nominal data.
The classification results differ between the two datasets. On Breast\_Cancer the $k$NN G-mean
of COFRS stays within $0.01$ of the best baseline at every rate. On Tic-Tac-Toe the
two-attribute reduct at $0\%$ is too small: its G-mean is $0.04$, against $0.56$ for Tol-RS, and
the gap closes only at $40\%$.
Figure~\ref{fig:diagnostics}b shows the effect of the granule threshold.
The mean reduct size grows from $1.5$ attributes at $\delta=0.05$ to $5.9$ at $\delta=0.30$.
At every value of $\delta$ it exceeds the mean granule floor $\lceil\delta|A|\rceil$ by at most $0.13$ attributes.
The mean $k$NN G-mean rises from $0.57$ to $0.68$ at $\delta=0.15$ and then varies by less than $0.04$.

\paragraph{Horse\_Colic.}
\label{sec:case-study}
Horse\_Colic has the highest missing rate ($24.35\%$, IR $= 4.05$). The tolerance-based methods keep
$9.4$--$16.8$ of its $22$ attributes and COFRS keeps $6.0$, while the $k$NN accuracy of every method
lies between $0.57$ and $0.63$.
The SVM G-mean is at most $0.07$ for \emph{all} methods on this dataset, probably because the
default cost-insensitive SVM predicts mostly the majority class~\cite{He2009}. It does not depend
on the selected attributes.

\paragraph{Running time.}
Table~S4 of the Supplementary material reports the time of the reduction step.
On the ten datasets with $n\le 961$, all methods finish in under three seconds per fold.
COFRS is among the slowest ($2.37$\,s on average, against $0.02$\,s for Zhou-RS), because it
evaluates $\gamma_{\mathrm{soft}}$ for every candidate removal in the backward loop.
On Hypothyroid ($n=3163$), COFRS-min, COFRS and Tol-RS$_\tau$ take $19$--$26$\,s per fold.

\subsection{Size-matched control}
\label{subsec:size_matched}

The granule-emptiness floor $|B^*|\geq\lceil\delta|A|\rceil$
(Remark~\ref{rem:self_membership}) raises a legitimate concern, since the search stops there in
most folds. A small reduct is therefore not by itself evidence that COFRS has selected \emph{good}
attributes, since $k$ \emph{arbitrary} attributes would have the same cardinality.
We therefore isolate the selection quality from the cardinality by a size-matched control.
For each dataset and fold we fix $k=|B^*_{\mathrm{COFRS}}|$ and compare COFRS with two kinds of
control. The first is a \emph{random} subset of $k$ attributes, averaged over $30$ independent draws.
The second ranks the attributes by each baseline's own criterion and keeps the top $k$.
For Tol-RS, Tol-RS$_\tau$ and COFRS-min the ranking is a forward greedy search over all of $A$
that stops at $k$ attributes. It maximises the tolerance dependency for the first two, and the
soft consistency of the $\min$-aggregated relation without a granule threshold ($\delta=0$) for
COFRS-min. For Zhou-RS it lists the kept attributes by increasing number of undiscerned pairs,
then the deleted ones in reverse order of deletion.
All subsets are evaluated with the same 5-NN classifier on the same 5-fold splits.

\begin{table}[pos=htbp]
\centering
\caption{Size-matched control: COFRS reduct of size $k=|B^*_{\mathrm{COFRS}}|$ versus a random
subset of size $k$ (mean over 30 draws), 5-NN, 5-fold CV.}
\label{tab:size_matched}
\footnotesize
\begin{tabular*}{\linewidth}{@{\extracolsep{\fill}}lrrrrr@{}}
\toprule
 & & \multicolumn{2}{c}{G-mean} & \multicolumn{2}{c}{Accuracy}\\
\cmidrule(lr){3-4}\cmidrule(lr){5-6}
Dataset & $k$ & COFRS & Random & COFRS & Random\\
\midrule
Hepatitis        & 4.8 & \textbf{0.529} & 0.462 & 0.774 & 0.808 \\
Horse\_Colic & 6.0 & \textbf{0.213} & 0.128 & \textbf{0.619} & 0.596 \\
Cleveland        & 3.4 & \textbf{0.787} & 0.696 & \textbf{0.796} & 0.706 \\
Pima             & 3.0 & \textbf{0.693} & 0.598 & \textbf{0.738} & 0.672 \\
Credit           & 2.6 & \textbf{0.834} & 0.571 & \textbf{0.838} & 0.634 \\
Automobile       & 4.8 & 0.833 & 0.838 & 0.836 & 0.842 \\
Mammographic     & 1.6 & \textbf{0.778} & 0.684 & \textbf{0.789} & 0.716 \\
Cylinder\_Bands  & 8.8 & 0.624 & 0.633 & 0.654 & 0.665 \\
Congressional    & 1.0 & \textbf{0.958} & 0.404 & \textbf{0.956} & 0.693 \\
Heart\_Hungarian & 3.6 & \textbf{0.737} & 0.565 & \textbf{0.779} & 0.711 \\
Hypothyroid      & 4.2 & \textbf{0.807} & 0.576 & \textbf{0.971} & 0.648 \\
\midrule
\textbf{Mean} &  & \textbf{0.708} & 0.560 & \textbf{0.795} & 0.699 \\
\bottomrule
\end{tabular*}
\end{table}

Table~\ref{tab:size_matched} shows the result. At \emph{identical} cardinality the
COFRS reduct outperforms an equal-size random subset on G-mean on 9 of the 11 datasets
(mean $0.708$ vs.\ $0.560$). The advantage is significant under a Wilcoxon signed-rank
test across datasets, with $p=0.005$ for G-mean and F1 and $p=0.019$ for accuracy. The gap is
largest on Congressional, Credit and Hypothyroid. The conservative-overlap criterion therefore
selects informative attributes and does not merely hit a cardinality target.

Table~\ref{tab:size_matched_wilcoxon} compares COFRS with the size-matched rankings of the baselines,
with Holm's correction over the five controls. COFRS is significantly better on G-mean than random
subsets and than the Tol-RS ranking (Holm-adjusted $p=0.024$ and $p=0.027$). It is better than the
Tol-RS$_\tau$ ranking on 9 of the 11 datasets and than the Zhou-RS ranking on 7 of the 10 datasets
where the two differ, but neither difference is significant.
It does not differ from the COFRS-min ranking ($p=0.63$), which uses the same soft consistency
with the $\min$ aggregation. The advantage of COFRS over COFRS-min therefore lies in the small $k$
itself: the COFRS-min reduct averages $10.3$ attributes (Table~\ref{tab:reduct_size}).

\begin{table}[pos=htbp]
\centering
\caption{Size-matched comparison against controls (G-mean, Wilcoxon signed-rank across 11
datasets). Each baseline keeps the top $k=|B^*_{\mathrm{COFRS}}|$ attributes of its own
ranking; ``Random'' picks $k$ attributes at random (mean over 30
draws). $p$: exact two-sided signed-rank test, zero differences dropped; $^{*}$ marks Holm-adjusted $p<0.05$.}
\label{tab:size_matched_wilcoxon}
\footnotesize
\begin{tabular*}{\linewidth}{@{\extracolsep{\fill}}lrrrrrr@{}}
\toprule
Control & COFRS mean & Ctrl mean & Wins/Losses & $W$ & $p$ & Holm $p$\\
\midrule
Random & 0.708 & 0.560 & 9/2 & 3.0 & 0.0049 & 0.0244$^{*}$\\
Tol-RS & 0.708 & 0.584 & 10/1 & 4.0 & 0.0068 & 0.0273$^{*}$\\
Tol-RS$_\tau$ & 0.708 & 0.627 & 9/2 & 11.0 & 0.0537 & 0.1611\\
Zhou-RS & 0.708 & 0.695 & 7/3 & 15.0 & 0.2324 & 0.4648\\
COFRS-min & 0.708 & 0.699 & 5/5 & 22.0 & 0.6250 & 0.6250\\
\bottomrule
\end{tabular*}
\end{table}

\subsection{Comparison with the Jensen--Shen selector and NETR}
\label{subsec:faithful_baselines}

The comparison of Section~\ref{subsec:main_results} contains no genuine fuzzy-rough selector
and omits the neighbourhood-equivalence method of
Wu et al.~\cite{Wu2024}. To test whether COFRS's
compactness depends on the choice of competitor, we implement both.
We evaluate them under the identical protocol (same eleven incomplete UCI
datasets, same outer $5$-fold stratified split with seed $42$, same $k$NN with
$k{=}5$).

\emph{Jensen--Shen FRFS.} We implement the fuzzy positive-region QuickReduct.
It is a forward greedy search. At each step it adds the attribute that maximises the fuzzy
dependency $\gamma_B=|\underline{\mathrm{POS}}_B(D)|/|U|$, where the fuzzy lower
approximation aggregates per-attribute similarities with the $\min$ t-norm.
It stops when $\gamma_B$ reaches the full-attribute dependency.
Jensen and Shen do not treat missing values. We set $\mu_a(x,y)=0$ when either value is missing,
as in COFRS. Under the $\min$ t-norm an object with a missing value on $B$ is then similar to no
object, so it falls in the lower approximation of its class without evidence. This convention
favours small FRFS reducts on data with many missing values. \emph{Wu 2024 NETR.} Two objects are neighbourhood-tolerant if the Euclidean distance over the
attributes present in both is at most $\delta$; they are related by the neighbourhood equivalence
tolerance relation if their neighbourhood-tolerance classes coincide. The search is the forward
algorithm of Wu et al., which adds the attribute with the largest dependency gain until the dependency
of the full set is reached. Wu et al.\ set $\delta$ per dataset; we select
$\delta\in\{0.1,\dots,0.6\}$ on each training fold by inner $5$-fold CV on G-mean, as for COFRS.
Categorical attributes, which Wu et al.\ do not treat, contribute distance $0$ if equal and $1$ otherwise.

\begin{table}[pos=htbp]
\centering
\caption{Comparison with the Jensen--Shen selector (FRFS) and the neighbourhood equivalence tolerance relation of Wu et al.~\cite{Wu2024} (NETR); $k$NN ($k{=}5$), 5-fold CV.}
\label{tab:m6_baselines}
\scriptsize
\begin{tabular*}{\linewidth}{@{\extracolsep{\fill}}lrrrrrrrr@{}}
\toprule
 & \multicolumn{2}{c}{COFRS} & \multicolumn{3}{c}{FRFS (Jensen--Shen)} & \multicolumn{3}{c}{NETR (Wu 2024)}\\
\cmidrule(lr){2-3}\cmidrule(lr){4-6}\cmidrule(lr){7-9}
Dataset & $|R|$ & Acc & $|R|$ & Acc & G-mean & $|R|$ & Acc & G-mean\\
\midrule
Hepatitis & \textbf{4.8} & 0.774 & 12.4 & 0.845 & 0.661 & 12.2 & 0.787 & 0.603\\
Horse\_Colic & 6.0 & 0.619 & \textbf{4.8} & 0.602 & 0.291 & \textbf{4.8} & 0.629 & 0.187\\
Cleveland & \textbf{3.4} & 0.796 & 13.0 & 0.818 & 0.814 & 12.6 & 0.822 & 0.817\\
Pima & \textbf{3.0} & 0.738 & 8.0 & 0.733 & 0.693 & 5.2 & 0.716 & 0.653\\
Credit & \textbf{2.6} & 0.838 & 14.0 & 0.814 & 0.811 & 14.6 & 0.804 & 0.800\\
Automobile & \textbf{4.8} & 0.836 & 9.8 & 0.836 & 0.835 & 16.4 & 0.846 & 0.842\\
Mammographic & \textbf{1.6} & 0.789 & 5.0 & 0.781 & 0.779 & 4.6 & 0.779 & 0.777\\
Cylinder\_Bands & 8.8 & 0.654 & \textbf{7.8} & 0.640 & 0.612 & 21.8 & 0.651 & 0.619\\
Congressional & \textbf{1.0} & 0.956 & 8.0 & 0.938 & 0.934 & 12.8 & 0.936 & 0.936\\
Heart\_Hungarian & 3.6 & 0.779 & \textbf{1.8} & 0.660 & 0.135 & 12.6 & 0.820 & 0.782\\
Hypothyroid & 4.2 & 0.971 & \textbf{2.8} & 0.815 & 0.686 & 20.4 & 0.975 & 0.712\\
\midrule
\textbf{Mean} & \textbf{3.98} & 0.795 & 7.95 & 0.771 & 0.659 & 12.55 & 0.797 & 0.703\\
\bottomrule
\end{tabular*}
\end{table}

Both methods select more attributes than COFRS. FRFS keeps $7.95$ attributes on average and
NETR $12.55$, against $3.98$ for COFRS.
COFRS is smaller on $7$ of $11$ datasets against FRFS and on $10$ of $11$ against NETR
(Wilcoxon signed-rank on reduct size, unadjusted $p=0.042$ and $p=0.002$).
The advantage over FRFS is modest: with a Bonferroni correction over the two comparisons,
its $p$ becomes $0.084$.
The $k$NN accuracy does not differ significantly ($p=0.18$ against FRFS and $p=0.77$ against NETR).

\subsection{Statistical Analysis}
\label{subsec:statistical_analysis}

\begin{figure}[pos=htbp]
\centering
\begin{subfigure}[t]{0.48\textwidth}
\centering
\resizebox{\linewidth}{!}{
\begin{tikzpicture}[x=1cm, y=1cm, every node/.style={inner sep=0pt}]
  \cdAxis{1}{4}
  \cdCDscale{1.84}{1.8392172247997245}
  \cdTitle{$|R|$}
  \cdLeft{1.318}{COFRS}{1}
  \cdLeft{2.045}{Tol-RS}{2}
  \cdRight{3.636}{Tol-RS$_\tau$}{1}{0.0}
  \cdRight{3.773}{COFRS-min}{2}{0.0}
  \cdRight{4.227}{Zhou-RS}{3}{0.0}
  \cdClique{1.318}{2.045}{0.2}{0.0}
  \cdClique{2.045}{3.773}{0.32}{0.0}
  \cdClique{3.636}{4.227}{0.44}{0.0}
\end{tikzpicture}
}
\caption{$|R|$}
\label{fig:cd_reduct}
\end{subfigure}
\hfill
\begin{subfigure}[t]{0.48\textwidth}
\centering
\resizebox{\linewidth}{!}{
\begin{tikzpicture}[x=1cm, y=1cm, every node/.style={inner sep=0pt}]
  \cdAxis{1}{5}
  \cdCDscale{2.27}{2.273513500369777}
  \cdTitle{$k$NN accuracy}
  \cdLeft{2.909}{COFRS-min}{1}
  \cdLeft{3.364}{COFRS}{2}
  \cdRight{3.455}{Full}{1}{0.0}
  \cdRight{3.682}{Tol-RS$_\tau$}{2}{0.0}
  \cdRight{3.727}{Zhou-RS}{3}{0.0}
  \cdRight{3.864}{Tol-RS}{4}{0.0}
  \cdClique{2.909}{3.864}{0.2}{0.0}
\end{tikzpicture}
}
\caption{kNN accuracy}
\label{fig:cd_knn_accuracy}
\end{subfigure}
\par\medskip
\begin{subfigure}[t]{0.48\textwidth}
\centering
\resizebox{\linewidth}{!}{
\begin{tikzpicture}[x=1cm, y=1cm, every node/.style={inner sep=0pt}]
  \cdAxis{1}{5}
  \cdCDscale{2.27}{2.273513500369777}
  \cdTitle{SVM accuracy}
  \cdLeft{2.955}{Tol-RS$_\tau$}{1}
  \cdLeft{3.045}{Full}{2}
  \cdLeft{3.136}{Zhou-RS}{3}
  \cdLeft{3.182}{COFRS-min}{4}
  \cdRight{3.909}{Tol-RS}{1}{0.0}
  \cdRight{4.773}{COFRS}{2}{0.0}
  \cdClique{2.955}{4.773}{0.2}{0.0}
\end{tikzpicture}
}
\caption{SVM accuracy}
\label{fig:cd_svm_accuracy}
\end{subfigure}
\hfill
\begin{subfigure}[t]{0.48\textwidth}
\centering
\resizebox{\linewidth}{!}{
\begin{tikzpicture}[x=1cm, y=1cm, every node/.style={inner sep=0pt}]
  \cdAxis{1}{5}
  \cdCDscale{1.84}{1.8392172247997245}
  \cdTitle{Runtime (s)}
  \cdLeft{1.000}{Zhou-RS}{1}
  \cdLeft{2.364}{Tol-RS}{2}
  \cdLeft{2.636}{Tol-RS$_\tau$}{3}
  \cdRight{4.273}{COFRS-min}{1}{0.0}
  \cdRight{4.727}{COFRS}{2}{0.0}
  \cdClique{1.000}{2.636}{0.2}{0.0}
  \cdClique{2.636}{4.273}{0.32}{0.0}
  \cdClique{4.273}{4.727}{0.44}{0.0}
\end{tikzpicture}
}
\caption{Runtime (s)}
\label{fig:cd_runtime}
\end{subfigure}
\caption{Critical Difference (CD) diagrams for the four primary metrics (Nemenyi post-hoc; $\alpha{=}0.05$, $N{=}11$). Lower rank is better; thick bars connect non-significant cliques. F1 and G-mean ranks appear in Table~\ref{tab:avg_ranks}.}
\label{fig:cd_all}
\end{figure}
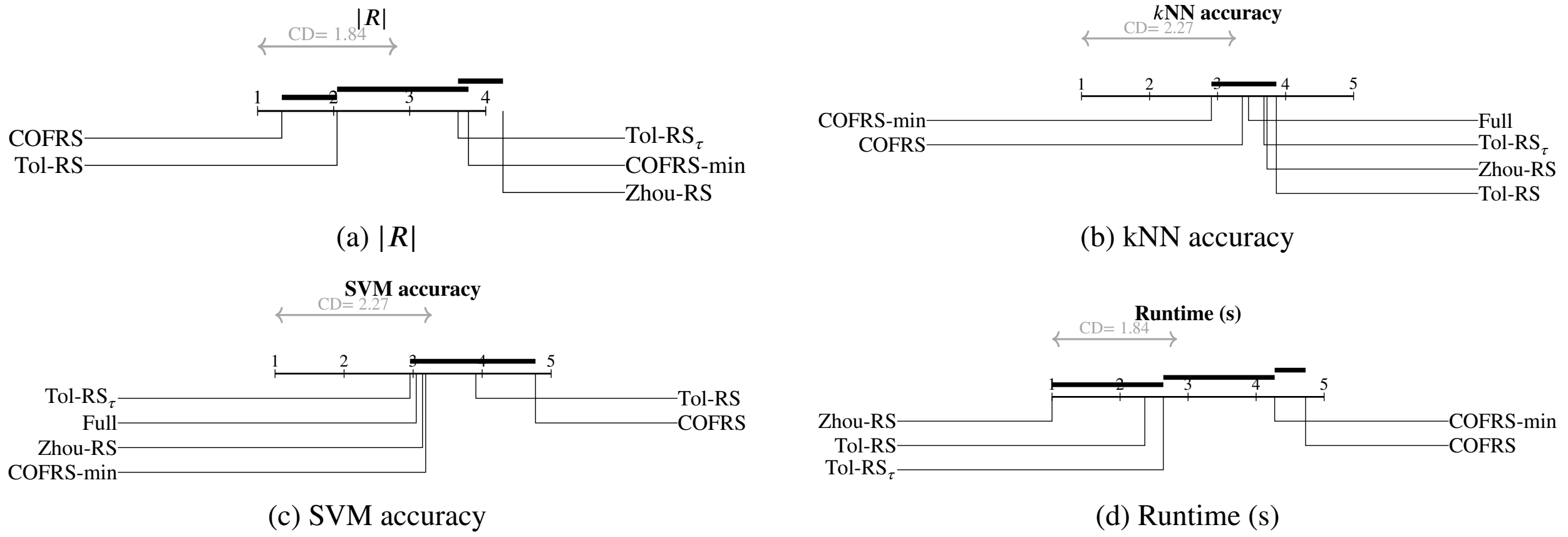

Table~\ref{tab:avg_ranks} summarises average ranks; Figure~\ref{fig:cd_all} shows the
corresponding Critical Difference diagrams (Friedman/Nemenyi tests below).

\begin{table}[pos=htbp]
\begin{threeparttable}
\centering
\footnotesize
\caption{Average ranks over 11 datasets (lower is better). \textbf{Bold}: best per row.}
\label{tab:avg_ranks}
\setlength{\tabcolsep}{3pt}
\begin{tabular*}{\textwidth}{@{\extracolsep{\fill}}lcccccc@{}}
\toprule
Metric & Tol-RS & Tol-RS$_\tau$ & Zhou-RS & \fRFS & COFRS & Full \\
\midrule
$|R|$ & 2.05 & 3.64 & 4.23 & 3.77 & \textbf{1.32} & --- \\
$k$NN Acc. & 3.86 & 3.68 & 3.73 & \textbf{2.91} & 3.36 & 3.45 \\
$k$NN F1 & 4.05 & 3.14 & 4.09 & \textbf{3.00} & 3.18 & 3.55 \\
$k$NN G-mean & 4.14 & 3.14 & 3.73 & \textbf{2.73} & 4.00 & 3.27 \\
SVM Acc. & 3.91 & \textbf{2.95} & 3.14 & 3.18 & 4.77 & 3.05 \\
SVM F1 & 4.05 & 3.27 & 2.91 & 3.32 & 4.64 & \textbf{2.82} \\
SVM G-mean & 4.23 & \textbf{2.91} & 3.00 & 3.14 & 4.73 & 3.00 \\
Runtime (s) & 2.36 & 2.64 & \textbf{1.00} & 4.27 & 4.73 & --- \\
\bottomrule
\end{tabular*}
\begin{tablenotes}[flushleft]
\footnotesize
\item Reduct size $|R|$ and runtime rank only the five reduction methods; the full set is excluded.
\end{tablenotes}
\end{threeparttable}
\end{table}

We ran Friedman tests~\cite{Demsar2006,Friedman1940} on each metric over the 11 datasets,
with the methods of Tables~\ref{tab:reduct_size}--\ref{tab:gmean}.
Reduct size differs significantly across the five reduction methods ($\chi^2 = 30.09$,
$p < 10^{-5}$), and so does reduction runtime ($\chi^2 = 40.22$, $p < 10^{-6}$).
The six-method tests on classification, which include the full set, are not significant with
$k$NN ($p \geq 0.39$ for accuracy, F1 and G-mean).
With the RBF-SVM the test is significant for G-mean ($\chi^2 = 11.23$, $p = 0.047$) and narrowly
misses the $0.05$ level for accuracy and F1 ($p = 0.088$ and $p = 0.099$).
COFRS has the worst average SVM rank on all three metrics, but the Nemenyi test (CD $=2.27$)
separates no pair: the largest rank gap is $1.82$.

Wilcoxon signed-rank tests~\cite{Wilcoxon1945} on reduct size, with Holm--Bonferroni
correction over the four pairwise comparisons, confirm Section~\ref{subsec:main_results}.
COFRS is smaller on all 11 datasets than Tol-RS$_\tau$, Zhou-RS and COFRS-min (raw $p=0.001$,
Holm-adjusted $p=0.004$, rank-biserial $r=1.00$), and the comparison with Tol-RS is not
significant (raw $p = 0.14$, $r = 0.55$).
The low SVM ranks of COFRS are not explained by Hypothyroid alone, where its SVM G-mean is low and
unstable ($0.40\pm0.30$): its SVM accuracy is below Tol-RS$_\tau$ by more than $0.01$ on 5 of 11
datasets and above it on one.

In the Nemenyi diagrams of Figure~\ref{fig:cd_all}, COFRS has the best rank on reduct size ($1.32$,
CD $=1.84$) and shares a clique only with Tol-RS ($2.05$); its gap to Tol-RS$_\tau$ is $2.32$.
On runtime Zhou-RS has the best rank and COFRS the last.
For classification (six methods, CD $=2.27$), every panel is a single clique of all six methods.

\section{Discussion}
\label{sec:discussion}

\subsection{Interpreting the results}
\label{subsec:interpretation}

On data with many missing values, wildcard matching makes many objects look alike on small
subsets, so more attributes are needed to separate the classes.
On Horse\_Colic ($24.35\%$ missing) the tolerance-based methods keep $9.4$--$16.8$ attributes,
and COFRS keeps $6.0$.
On Pima, where the zeros are recoded as missing, COFRS keeps $3$ of $8$ attributes with a $k$NN
accuracy of $0.74$, against $8$ attributes and $0.73$ for Tol-RS$_\tau$.
Under injected missingness the COFRS reduct grows more slowly than the baseline reducts on
numeric data and faster on nominal data (Figure~\ref{fig:diagnostics}a).

The reduct size of COFRS follows the threshold $\delta$. In most folds the search stops at the
granule floor $\lceil\delta|A|\rceil$ (Section~\ref{subsec:finding_reduct}), and the baselines have no
corresponding parameter. A comparison of sizes alone therefore
overstates what the criterion contributes. The size-matched control of
Section~\ref{subsec:size_matched} is the more informative test. At equal size, COFRS keeps better
attributes than random subsets and the Tol-RS ranking; against the other rankings the difference is
not significant after Holm's correction.

COFRS is less stable across folds than the baselines (Table~S5 of the Supplementary material).
When a reduct has only a few attributes, replacing one of them on another fold changes a large
share of the set. The size also follows the per-fold choice of $\delta$, which on Hepatitis ranges
from $0.05$ to $0.30$.

With $k$NN we find no significant accuracy difference between the methods
(Section~\ref{subsec:statistical_analysis}), although a non-significant test does not show
that they are equivalent. Even without reduction, $k$NN reaches a mean accuracy of only $0.797$ on
these datasets, so a selector has little room to raise it. With the RBF-SVM COFRS has the lowest
ranks: its accuracy is below that of Tol-RS$_\tau$ by more than $0.01$ on $5$ of $11$
datasets and above it on one. The SVM with all attributes reaches $0.814$, but it uses $18.2$
attributes on average against $4.0$ for COFRS. In our experiments the compact reducts therefore
suit $k$NN better than the SVM.

The $\delta$-granule $[x]_\delta$ is a variable-precision granule
\cite{Pedrycz2013,Yao2001,Yao2016}. For $\delta\to 0$ it becomes $U$, and a larger $\delta$ gives
smaller granules. An object whose self-similarity $|\Vobs{x}|/|A|$ is below $\delta$ has an empty
granule (Remark~\ref{rem:self_membership}).

\subsection{Limits of the conservative bias}
\label{subsec:when_not}

COFRS charges every missing value against the fixed denominator $|A|$. This assumes that a missing
entry gives no evidence of similarity. The assumption fits an absent measurement, such as a
laboratory test that was not ordered or a sensor that was offline. Horse\_Colic and Pima are plausibly of this kind.

The assumption fits less well in three cases.
First, a missing value can be an informative level, for example ``not applicable'' for a subgroup.
Penalising it discards this signal, and such values should be recoded as an explicit category before
reduction.
Second, on purely categorical data $\RCO$ reduces to a count of matching attributes. We did not
compare with dedicated categorical methods such as the fuzzy rough reduction of He et al.~\cite{He2022}, which may
do better. Congressional is the only purely categorical benchmark; COFRS gives a $1.0$-attribute
reduct there with the highest $k$NN accuracy, but one dataset is not enough to decide.
Third, when few values are missing the penalty has little to act on, and $\RCO$ approaches the
average-aggregation fuzzy rough similarity (Remark~\ref{prop:frs_avg_relation}). We expect
little advantage in that regime but did not test it: the lowest missing rate among the benchmarks is
$0.15\%$ (Cleveland).

The evaluation has three limits. It uses one run of stratified 5-fold cross-validation (seed 42)
on 11 datasets of moderate size, so small differences are not resolved. The baselines Tol-RS$_\tau$,
Zhou-RS and COFRS-min use fixed thresholds (for Zhou-RS only on continuous attributes), whereas COFRS tunes $\delta$ on
G-mean in each fold. The compactness therefore compares a tuned method with untuned ones
(Remark~\ref{rem:delta_leakage}). Varying the baseline thresholds does not close the gap
(Section~\ref{subsec:main_results}). The comparison with two further baselines is limited to
$k$NN (Section~\ref{subsec:faithful_baselines}).

\section{Conclusion}
\label{sec:conclusion}

This paper developed COFRS, a fuzzy rough set model for attribute reduction in incomplete
mixed information systems. Its similarity $\RCO$ charges every missing value against the
fixed number of attributes $|A|$, so a missing value is never treated as a wildcard.
The soft dependency $\gamma_{\mathrm{soft}}$ built on it is not monotone in the attribute set, and a
greedy backward search with a tolerance therefore replaces the classical reduct.

In the experiments on 11 incomplete UCI benchmarks, $\delta$ largely fixed the number of selected
attributes, because the search stops at the granule floor $\lceil\delta|A|\rceil$ in most folds.
With that number fixed, COFRS still beat random subsets and the Tol-RS ranking on $k$NN G-mean;
against the other rankings the difference was not significant.
Its reducts were significantly smaller than those of Tol-RS$_\tau$ and Zhou-RS but not than those
of Tolerance-RS. With $k$NN its accuracy did not differ significantly from the baselines, and with
the RBF-SVM it had the lowest ranks. Its reducts were also less stable across folds.

The method has three main limits. The greedy search finds a locally compact reduct and not a
global maximiser of $\gamma_{\mathrm{soft}}$, and its cost grows with $|U|^2|A|^2$.
The sufficient condition for monotonicity (Proposition~\ref{prop:gamma_mono_cond}) is not tight.
The evaluation rests on one run of 5-fold cross-validation, and the neighbourhood method of
Li et al.~\cite{Li2023} was not compared.

Three directions follow. A rule that sets $\delta$ from the data would remove the nested search and
make the reduct size less dependent on tuning.
Treating $\delta$ as a resolution parameter of a
hierarchical granular model~\cite{Pedrycz2013,Yao2016} is one way to obtain it.
Class-weighted learning may address the weak SVM results on skewed data such as Horse\_Colic and
Hypothyroid~\cite{He2009,Fernandez2018}. An incremental update of $\RCO$ and
$\gamma_{\mathrm{soft}}$ would suit evolving information systems.

\section*{Declaration of competing interest}
The authors declare that they have no known competing financial interests or personal relationships that could have appeared to influence the work reported in this paper.

\section*{Data availability}
All datasets are publicly available from the UCI Machine Learning Repository~\cite{UCI2023}. The COFRS implementation, experiment scripts and result files are available at \url{https://github.com/ttdaiuneti/cofrs}. Every table in this paper is generated from these result files.

\section*{Funding}
This research received no specific grant from any funding agency in the public, commercial, or not-for-profit sectors.

\section*{CRediT authorship contribution statement}
Trung Hieu Vuong: Conceptualization, Methodology, Software, Validation, Formal analysis, Investigation, Data curation, Visualization, Writing -- original draft, Writing -- review \& editing.

Thanh Dai Tran: Conceptualization, Methodology, Software, Validation, Formal analysis, Investigation, Data curation, Visualization, Writing -- original draft, Writing -- review \& editing.

Van Thien Nguyen: Writing -- review \& editing.

Nhu Son Nguyen: Writing -- review \& editing.

\bibliographystyle{unsrtnat}
\bibliography{references}

\clearpage
\includepdf[pages=-]{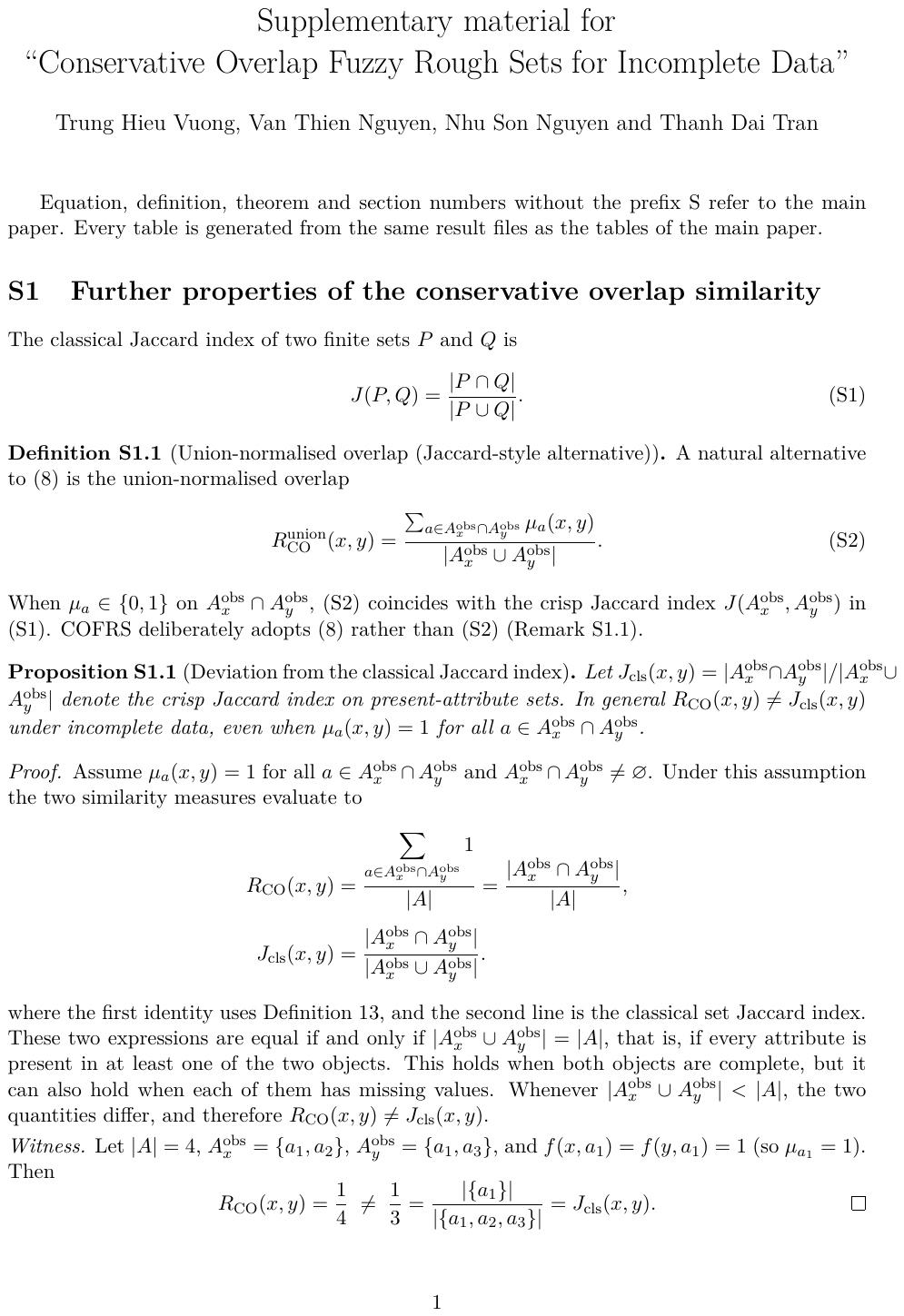}
\end{document}